\PassOptionsToPackage{table}{xcolor}
\documentclass[lettersize,journal]{IEEEtran}

\usepackage{booktabs}
\usepackage{tabularx}
\usepackage{multirow}
\usepackage{amsmath}
\usepackage{tcolorbox}
\usepackage{enumitem}
\usepackage{listings}
\usepackage[table]{xcolor}
\usepackage{amssymb}
\usepackage{hyperref}

\usepackage{amsmath,amsfonts}
\usepackage{algorithmic}
\usepackage{algorithm}
\usepackage{array}
\usepackage[caption=false,font=normalsize,labelfont=sf,textfont=sf]{subfig}
\usepackage{textcomp}
\usepackage{stfloats}
\usepackage{url}
\usepackage{verbatim}
\usepackage{graphicx}
\usepackage{cite}
\lstdefinestyle{pycode}{
  language=Python,
  basicstyle=\ttfamily\scriptsize,
  keywordstyle=\relax,          commentstyle=\relax,
  stringstyle=\relax,
  identifierstyle=\ttfamily,
  numbers=left, numberstyle=\tiny, numbersep=6pt,
  showstringspaces=false, upquote=true,
  breaklines=true, breakatwhitespace=true, columns=fullflexible,
  frame=single, framesep=3pt, rulecolor=\color{black}
}

\newcommand{\arxivcopyrightnotice}{%
\begin{center}
\begin{minipage}{0.96\linewidth}
\footnotesize
\textcopyright~2026 IEEE. Personal use of this material is permitted. Permission from IEEE must be obtained for all other uses, in any current or future media, including reprinting/republishing this material for advertising or promotional purposes, creating new collective works, for resale or redistribution to servers or lists, or reuse of any copyrighted component of this work in other works.
\end{minipage}
\end{center}
\vspace{-0.4em}
}

\begin{document}

\title{TQA-Bench: Evaluating LLMs for Multi-Table Question Answering}

\author{
Zipeng Qiu$^*$, Chenyue Li$^*$, You Peng, Guangxin He, Binhang Yuan, Chen Wang%
\thanks{Zipeng Qiu, Chenyue Li, You Peng, Guangxin He, and Binhang Yuan are with the Hong Kong University of Science and Technology (HKUST), Hong Kong, China. Chen Wang is with Tsinghua University, Beijing, China.}%
\thanks{Corresponding to Binhang Yuan (\textit{biyuan@ust.hk}) and Chen Wang (\textit{ wang\_chen@tsinghua.edu.cn}).}%
\thanks{Code and processed data are available at \href{https://github.com/Relaxed-System-Lab/TQA-Bench}{GitHub: Relaxed-System-Lab/TQA-Bench}.}
}

\maketitle
\arxivcopyrightnotice

\begin{abstract}
The advance of large language models (LLMs) has unlocked great opportunities in complex multi-modal data management tasks, particularly in question answering (QA) over complicated multi-table relational data. Despite significant progress, systematically evaluating LLMs on multi-table QA remains a critical challenge due to the inherent complexity of analyzing the modality of relational data structures and the potentially large scale of serialized tabular data. Existing benchmarks primarily focus on single-table QA, failing to capture the intricacies of connections across multiple relational tables, as required in real-world domains such as finance, healthcare, and e-commerce. 
We present TQA-Bench, a long-context analytical multi-table QA benchmark derived from real-world public datasets, with a flexible sampling mechanism that varies context length (8K--64K tokens) and symbolic extensions for assessing reasoning beyond retrieval and pattern matching.
We systematically evaluate a set of LLMs spanning model scales from 2 billion to 671 billion parameters.
Our extensive experiments reveal critical insights into the performance of LLMs in multi-table QA, highlighting both challenges and opportunities for advancing their application in complex, data-driven environments.
\end{abstract}


\section{Introduction}

Large language models (LLMs)~\cite{jin2022survey} have unlocked unprecedented opportunities in complex information-processing tasks~\cite{biswal2024text2sql,chentablerag,patel2024lotus,wornow2024automating}. While modalities such as unstructured text, vision, and audio have been widely explored in recent LLM research, structured relational data (e.g., relational tables in databases) are a high-utility yet relatively underexplored modality. This structured modality is particularly crucial for question answering (QA) tasks over real-world data~\cite{chentabfact,gu2022pasta,pal2023multitabqa, zhang2024reactable,zhu2024autotqa,he2024text2analysis}, where an LLM must interpret a natural language query and reason across multiple interrelated tables to generate the correct answer. Despite these advancements, systematically evaluating LLM performance on multi-table QA remains a significant challenge due to the task’s inherent complexity. Multi-table QA compels an LLM to engage in complex reasoning over structured relational data: the LLMs must extract and jointly analyze information from multiple interconnected tables, often contending with heterogeneous schemas and very large serialized inputs. As LLMs continue to demonstrate remarkable capabilities across various data management applications~\cite{patel2024lotus,wornow2024automating,madden2024databases,jiang2024siriusbi}, there is an urgent need for \textit{a comprehensive understanding of LLMs' performance in tackling the complexities of multi-table QA}  --- to assess how LLMs can reason over this structured data modality.

Systematically evaluating and understanding the performance of LLMs on multi-table QA is a crucial step toward unlocking their full potential for data management and business intelligence~\cite{chentabfact,lei2023tableqakit,wu2024tablebench}. Structured relational data is pervasive as a high-utility modality across domains~\cite{tatqa}, healthcare~\cite{zhu2019hierarchical}, and e-commerce~\cite{gao2021meaningful}. Real-world tasks often require the processing, retrieving, and analyzing of multiple tables to support data-driven decision making. However, there is \textit{a significant gap} between the existing Table QA benchmarks~\cite{chentabfact,lei2023tableqakit,wu2024tablebench} and the practical demands of applications that operate on real-world tabular data modality. We believe that addressing this disparity is essential to bridge the gap and advance the utility of the most advanced LLMs in complex, multi-modal data-centric environments.

We list the current table QA benchmarks in TABLE~\ref{tab:tableqabenchmarkinfo}, and summarize the challenges of constructing a practical table QA benchmark from three key aspects. \underline{First}, most existing Table QA benchmarks are designed based on single-table contexts~\cite{wikitablequestion,sqa,fetaqa,hybridqa,ottqa,finqa,aitqa,hitab,wu2024tablebench,tatqa}, 
which fail to capture the complex relationship across multiple interconnected tables in real-world scenarios.
\underline{Second}, the input lengths in existing benchmarks are often mismatched to contemporary long-context LLMs: many datasets contain only tiny tables (hundreds to a few thousand tokens), under-utilizing models that can handle much longer contexts~\cite{dubey2024llama,gpt4o}; meanwhile, some multi-table benchmarks serialize entire databases into extremely long inputs (e.g., MMQA~\cite{wu2025mmqa}), which can exceed the context budgets of commonly used models and thus require truncation or retrieval, entangling context management with reasoning.
\underline{Third}, evaluating analytical abilities over a fixed set of tables and questions raises concerns regarding the reliability and generalizability of the benchmark results, as models may merely learn to exploit probabilistic patterns in the dataset rather than exhibit robust performance on genuinely complex multi-table queries~\cite{mirzadeh2024gsm}. Addressing these challenges is essential to creating a benchmark that accurately reflects the demands of multi-table QA in real-world settings.

\begin{table*}[htbp]
\centering
\caption{The information of the current table QA benchmarks. For prior work, Avg tokens is computed per serialized table; for TQA-Bench (ours), per serialized database instance.}
\begin{tabular}{l m{6cm} c} 
\hline
\textbf{Benchmark} & \textbf{Tabular Data Source} & \textbf{Avg Tokens}\\
\hline
\rowcolor{lightgray}
\multicolumn{3}{c}{Single-Table QA} \\
\hline
WikiTableQuestions~\cite{wikitablequestion} & Wikipedia &1175.05\\
SQA~\cite{sqa} & Wikipedia &554.02\\
FetaQA~\cite{fetaqa} & Wikipedia &499.06\\
HybridQA~\cite{hybridqa} & Wikipedia &601.07\\
OTT-QA~\cite{ottqa} & Wikipedia &559.61\\
FinQA~\cite{finqa} & FinTabNet~\cite{zheng2021global} &190.37\\
AIT-QA~\cite{aitqa} & SecGov~\cite{secgov} &499.53\\
Hitab~\cite{hitab} & Wikipedia, Statistical reports &792.31\\
TableBench~\cite{wu2024tablebench} & Wikipedia, FinTabNet, SecGov &655.46\\
TATQA~\cite{tatqa} & Real-world financial report &447.91\\
\hline
\rowcolor{lightgray}
\multicolumn{3}{c}{Multi-Table QA} \\
\hline
Open-WikiTable~\cite{openwikitable} & Wikipedia &685.97\\
Multihiertt~\cite{multihiertt} & FinTabNet &1470.48\\
TSQA~\cite{tsqa} & Chinese high school exams &410.31\\
MMQA~\cite{wu2025mmqa} & Spider~\cite{yu2018spider} &87504.50\\
\hline
Ours & BIRD~\cite{bird}, DataGov~\cite{datagov}, WorldBank~\cite{worldbank} & Scale from 8K to 64K \\
\hline
\end{tabular}
\label{tab:tableqabenchmarkinfo}
\vspace{-1.0em}
\end{table*}

To address these challenges, we propose a novel design for multi-table QA benchmarks. \underline{First}, going beyond the simplistic single-table setups commonly used in current benchmarks, we construct the benchmark by systematically collecting multi-table relational database instances from diverse public datasets, where these datasets are carefully curated to represent practical real-world scenarios incorporating varied table structures, relationships, and domains. \underline{Second}, we introduce a novel controlled sampling mechanism to create evaluation tasks with varying multi-table context lengths, ranging from 8K to 64K tokens --- this mechanism enables us to assess the scalability of LLM's context length when processing multiple relation tables of different sizes, a critical requirement for real-world applications where data volumes can vary significantly. Adjustable context length is important for evaluating LLM's performance in its token limit.
Meanwhile, we can mitigate the risk of contamination and facilitate the regeneration of new evaluation splits if contamination is suspected.
\underline{Third}, to further reinforce the benchmark results' reliability, we incorporate symbolic extensions~\cite{mirzadeh2024gsm} into the evaluation framework, where flexible augmentation is integrated to evaluate the LLM's inherited reasoning ability over multi-table relational data instead of probabilistic retrieving or pattern matching. Both the sampling and symbolic extension methods make our benchmark sufficiently reliable and updateable periodically. Comprehensively, we adopt a principled design to construct TQA-Bench, a multi-table QA benchmark, to evaluate LLM performance on complex real-world relational QA tasks. Our contributions can be summarized below:

\begin{itemize}[topsep=5pt, leftmargin=1em]
    \item \textbf{A comprehensive multi-table QA benchmark.} We construct a scalable new benchmark for multi-table QA that addresses the limitations of existing single-table benchmarks. Our benchmark incorporates varied relational data contexts by employing a controlled sampling mechanism to generate evaluation tasks with context lengths ranging from 8K to 64K tokens. To ensure the reliability of evaluation results, we integrate symbolic extensions into the question templates, accessing the essential capabilities of LLMs beyond simple data retrieval or pattern matching.

    \item \textbf{A wide range of LLM evaluation results.} We systematically evaluate both open-source and closed-source LLMs on our benchmark, where the open-source models span a range of scales, from 2 billion to 671 billion parameters. We provide a comprehensive assessment of LLMs over challenging real-world multi-table QA tasks.

    \item \textbf{Key observations and insights.} Our comprehensive evaluation yields the following key insights: (\underline{\textbf{i}}) \textit{Single- vs. multi-table performance}: LLMs consistently perform better in single-table settings, achieving up to $20\%$ higher accuracy compared to multi-table scenarios. This highlights the inadequacy of existing single-table benchmarks in capturing the complexity of real-world analytical tasks and underscores the need for dedicated multi-table QA evaluation. (\underline{\textbf{ii}}) \textit{Table serialization format}: The choice of serialization format significantly affects the model performance. Markdown outperforms CSV, JSON, and HTML across most LLMs and context lengths, providing a more LLM-friendly structure. (\underline{\textbf{iii}}) \textit{Model category and context sensitivity}: Instruction-tuned LLMs significantly outperform chat-oriented and domain-specific models, particularly under long-context settings. Reasoning LLMs (e.g., \textsc{DeepSeek-R1}) achieve strong performance, while some distilled variants often fail to handle longer contexts. Overall, increasing context length leads to consistent performance degradation, especially for aggregation and complex analytical tasks.
    (\underline{\textbf{iv}}) \textit{Sampling and symbolic extension}: Our symbolic extension and database sampling strategies enhance benchmark diversity and robustness, introduce a wider range of query patterns and difficulty levels, reduce variance in evaluation results, and enable consistent assessment across different database instances and question templates.
    (\underline{\textbf{v}}) \textit{Direct prompting vs. Text2SQL}: LLM-based Text2SQL methods demonstrate stable performance across context lengths and offer a complementary approach to direct prompting. However, they still struggle with complex analytical queries,
    due to challenges in composing semantically correct SQL, and actually underperform the strongest LLMs by direct-prompting.

\end{itemize}

\section{Benchmark Construction}
\label{sec:construct}

We consider \textit{multi-table QA} as the task of answering a single natural language question using tabular data from two or more distinct tables that are semantically related (e.g., through foreign-key relationships). The correct answer may require joining information across tables and possibly performing computations or analytics over the combined data.

The construction process of our multi-table QA benchmark is systematically divided into four key phases: data collection, relational data sampling, evaluation task definition, and question generation with controlled symbolic extensions, where each phase can be summarized as follows:

\begin{itemize} [topsep=5pt, leftmargin=1em]
    \item \textbf{Collect multi-table data} (\S\ref{sec:collect}). To ensure the diversity and representativeness of our benchmark, we collect a wide variety of large-scale relational databases. These databases serve as the foundation for multi-table QA tasks.

    \item \textbf{Relational data sampling} (\S\ref{sec:sample}). We design a controlled sampling methodology to create subsets of each table with varying serialized lengths. This approach ensures that the sampled data maintains the structural integrity and heterogeneity of the original datasets to evaluate LLM's performance under different context lengths.

    \item \textbf{Define evaluation task categories} (\S\ref{sec:task_cate}). We define three primary question categories, further divided into seven subcategories, inspired by those commonly found in traditional Table QA datasets. These categories are designed to capture a broad spectrum of question types, reflecting the diverse requirements of real-world multi-table QA tasks.

     \item \textbf{Generate question with symbolic extension} (\S\ref{sec:symbolic}). For each question category, we develop structured question templates that are augmented with symbolic extensions to assess analysis capabilities beyond simple retrieval. These templates are paired with Python-based answer generation, enabling the automated creation of benchmark questions and ensuring scalability and reliability in task evaluation.    
\end{itemize}

This structured and systematic process enables the creation of a scalable, diverse, and effective benchmark for evaluating LLM performance on complex multi-table QA tasks.

\subsection{Multi-Table Data Collection} 
\label{sec:collect}

Many existing Table QA datasets are based on tables from Wikipedia~\cite{wikitablequestion,sqa,openwikitable,hybridqa,ottqa,hitab,fetaqa}. However, many LLMs are pre‐trained on Wikipedia, which risks potential bias and contamination. Moreover, Wikipedia’s tables are typically short (only tens of rows) and do not adequately challenge models on comprehensive Table QA tasks. To ensure a rigorous evaluation with complex, unfamiliar tables, we deliberately excluded Wikipedia-derived data. Our data collection instead considers the following sources:

\begin{itemize} [topsep=5pt, leftmargin=1em]
\item \textsc{WorldBank.} We incorporated datasets from WorldBank~\cite{worldbank} to overcome Wikipedia’s limitations. WorldBank tables feature extensive rows and columns with simple yet meaningful foreign key relationships that generate actionable insights. Our analysis shows that these datasets often have long-context, multi-table structures—characteristics missing in existing benchmarks. We selected a WorldBank dataset~\cite{dasgupta2024revisiting} that fits our experimental setup and challenges LLMs in realistic, complex scenarios.

\item \textsc{DataGov.}
DataGov~\cite{datagov} offers a rich source of real-world tables. Its datasets comprise tables with numerous rows and columns and include basic foreign key relationships that support multi-table reasoning. For our benchmark, we chose two representative DataGov datasets: the Water Quality Data~\cite{water_quality_data} and Food Facility Inspections~\cite{food_facility_inspections}. These datasets were sampled and scaled to different target context lengths, enabling a wide range of experimental setups.

\item \textsc{BIRD}.
To complement the above, we included seven databases from BIRD~\cite{bird}, a benchmark originally designed for Text2SQL tasks. BIRD databases resemble real-world multi-table environments with complex foreign key relationships; however, many lack referential integrity. Since our sampling requires acyclic, valid foreign key graphs for meaningful queries, we excluded about half of BIRD’s databases, narrowing the selection to 20. From these, we carefully chose seven databases that balance semantic richness and manageable complexity, aligning with our benchmark’s objectives. The detailed information of the selected databases is shown in TABLE~\ref{tab:origin_table_info}.
\end{itemize}

\begin{table*}[htbp]
\centering
\caption{Structural and query-complexity statistics of the ten TQA-Bench databases.}
\begin{tabular}{l c c c c c c c} 
\hline
\textbf{Database Name} & \textbf{Source} & \textbf{Table Count} & \textbf{Average \#Columns} &\textbf{ Average Rows} & \textbf{Total Cells} & \textbf{Join Depth}\\
\hline
airline & BIRD & 3 & 10.67 & $2.37\times 10^5$ & $1.97\times 10^7$ &1.79\\
food\_inspection & BIRD & 3 & 8.33 & $2.21\times 10^4$ & $3.77\times 10^5$ &1.64\\
movie & BIRD & 3 & 9.00 & $2.55\times 10^3$ & $6.00\times 10^4$ &1.21\\
music\_tracker & BIRD & 2 & 5.00 & $1.19\times 10^5$ & $1.01\times 10^6$ &1.5\\
restaurant & BIRD & 3 & 4.00 & $6.43\times 10^3$ & $8.66\times 10^4$ &1.64\\
university & BIRD & 6 & 3.33 & $5.34\times 10^3$ & $1.29\times 10^5$ &1.71\\
cookbook & BIRD & 4 & 9.75 & $2.59\times 10^3$ & $7.97\times 10^4$ &1.43\\
food\_facility\_inspections & DataGov & 3 & 13.67 & $1.69\times 10^5$ & $4.82\times 10^6$ &1.64\\
water\_quality & DataGov & 4 & 9.75 & $1.64\times 10^6$ & $7.01\times 10^7$ &1.21\\
global\_biodiversity & WorldBank & 2 & 15.50 & $5.97\times 10^5$ & $1.85\times 10^7$ &1.71\\
\hline
Overall Average & - & 3.3 & 8.36 & $2.83\times 10^5$ & $1.15\times 10^7$ &1.55\\
\hline
\end{tabular}
\label{tab:origin_table_info}
\vspace{-1em}
\end{table*}

\subsection{Sampling to Variate Context-Length}
\label{sec:sample}

TABLE~\ref{tab:origin_table_info} shows that many selected databases include tables with over 100,000 rows --- far beyond the token limits of mainstream LLMs, which makes direct construction of a multi-table QA dataset impractical.
To address this challenge, we develop a sampling method to generate databases of varying context lengths, enabling scalable benchmarking across experimental setups under different computational constraints.
To make the target scales well-defined and comparable across serializations, we treat 8K/16K/32K/64K as \emph{token-budget labels} rather than exact token counts in every format. Concretely, we enforce the length constraint using the token range measured under a Markdown serialization, which serves as the reference format for sampling control (TABLE~\ref{tab:scale}).
Since other formats (e.g., CSV/HTML/JSON) may expand the same underlying content to different token counts, this Markdown-based criterion provides a consistent and reproducible definition of context-length settings.
This sampling process involves two primary steps: (\underline{\textbf{i}}) determine the topological order of the tables based on their foreign key relationships to ensure referential integrity is maintained during sampling; (\underline{\textbf{ii}}) perform the row sampling for each table to create new multi-table database instances from the original databases. These steps ensure that the structural and relational properties of the databases are preserved, even at a reduced scale, allowing for effective benchmarking under various conditions.

\begin{table}[htbp]
\centering
\caption{Context-length settings as \emph{token-budget labels}, instantiated by Markdown-token ranges used for sampling control.}

\resizebox{\columnwidth}{!}{
\begin{tabular}{l c c} 
 \hline
 \textbf{Context Length} & \textbf{Minimum Token Limit} & \textbf{Maximum Token Limit}\\
 \hline
 8K & 4000 & 6000 \\
 16K & 8000 & 12000 \\
 32K & 16000 & 24000 \\
 64K & 32000 & 48000 \\
 \hline
\end{tabular}
}
\label{tab:scale}
\end{table}

\noindent\textbf{Topological Sort.}
The first step in our sampling procedure, as outlined in Algorithm~\ref{alg:topo}, involves determining a topological order among tables to preserve referential integrity during sampling. Formally, for tables within a database, \textit{if table $T_i$ references table $T_j$, then $T_i$ must be sampled prior to $T_j$}. This ordering relies on the assumption that table reference relationships constitute a directed acyclic graph (DAG). While alternative approaches exist for handling cyclic dependencies among tables, such methods complicate precise control over the number of sampled rows per table. Controlling the row count is essential for generating databases with variable context lengths, as it directly affects benchmark scalability across diverse computational settings.

\begin{algorithm}[tb]
\caption{Topological Sort}\label{alg:topo}
\begin{algorithmic}
\footnotesize
\REQUIRE $D$: Database instance with tables $\{T_1, ..., T_n\}$ and foreign key dependencies
\ENSURE A topologically sorted list of tables or an indication of a cyclic dependency
\STATE Initialize an empty list $L \gets \varnothing$
\STATE Create a map $R$, where $R[T_i]$ is the count of incoming references (in-degree) for table $T_i$
\STATE Initialize a set $S \gets \{T_i \mid R[T_i] = 0\}$ containing all tables with zero in-degree
\WHILE{$S \neq \varnothing$}
    \STATE{\textcolor{black}{/* Pick any table with zero in-degree */}}
    \STATE Select and remove a table $T \in S$
    \STATE Append $T$ to $L$
    \FOR{each table $U$ referenced by $T$ (i.e., $T \to U$)}
        \STATE $R[U] \gets R[U] - 1$
        \IF{$R[U] = 0$}
            \STATE Add $U$ to $S$
        \ENDIF
    \ENDFOR
\ENDWHILE
\STATE{\textcolor{black}{/* Check for remaining edges indicating a cycle */}}
\IF{$\exists T_i \text{ such that } R[T_i] > 0$}
    \RETURN ``Cycle detected``
\ENDIF
\RETURN $L$
\end{algorithmic}
\end{algorithm}

\begin{algorithm}[tb]
\caption{Row Sampling with Referential Integrity}\label{alg:sampling}
\begin{algorithmic}
\footnotesize
\REQUIRE $D$: Database instance
\REQUIRE $k$: Number of rows to sample from tables without incoming references
\ENSURE Sampled subset of $D$ maintaining referential integrity
\STATE Initialize an empty list $L \gets \varnothing$
\STATE Compute a topological order $O \gets \textsc{TopologySort}(D)$
\FOR{each table $U \in O$}
    \IF{$U$ has no incoming references}
        \STATE{\textcolor{black}{/* Sample $k$ rows from $U$ and preserve row order */}}
        \STATE $T \gets \textsc{KeepOrderSample}(U, k)$
    \ELSE
        \STATE Initialize an empty map $M \gets \varnothing$
        \STATE{\textcolor{black}{/* Column $A_R$ in $R$ references column $A_U$ in $U$ */}}
        \FOR{each reference $R.A_R \to U.A_U$}
            \STATE{\textcolor{black}{/* Add referenced values of $A_R$ in $M[A_U]$ */}}
            \STATE $M[A_U] \gets M[A_U] \cup R[A_R]$
        \ENDFOR
        \STATE Initialize an empty set $T \gets \varnothing$
        \FOR{each row $r \in U$}
            \IF{any attribute $A$ of $r$ satisfies $r[A] \in M[A]$}
                \STATE $T \gets T \cup \{r\}$
            \ENDIF
        \ENDFOR
    \ENDIF
    \STATE $L \gets L \cup T$
\ENDFOR
\RETURN $L$
\end{algorithmic}
\end{algorithm}

\noindent\textbf{Row Sampling.}
The second step, detailed in Algorithm~\ref{alg:sampling}, involves sampling rows from the database while preserving referential integrity. Given a parameter $k$ to determine the sampled number of rows, the algorithm handles tables differently based on their reference dependencies. For tables without incoming references, an ordered sampling of $k$ rows is performed directly. For tables referenced by others, sampling is guided by the topological order of the tables. Rows are selected from the original table that match the referenced column values in the sampled tables, ensuring that all foreign key constraints are respected. To determine the token counts accurately for the sampled databases, we serialize the tables into Markdown format, include table names, and calculate token sizes using a tokenizer. By adjusting the parameter $k$, databases with varying token sizes are generated, approximating the desired context length through a binary search approach. For each target context length, ten database instances were sampled for each original database. The details of the sampled databases are summarized in TABLE~\ref{tab:sampled_table_info}.
Overall, the current benchmark covers a moderate multi-table setting, while the sampling procedure itself applies more broadly to DAG-structured relational schemas.

\begin{table*}[tb]
\centering
\caption{The detailed information on ten databases under four different context lengths.}

\begin{tabular}{ l|cccc|cccc  }
 \hline
 & \multicolumn{4}{|c|}{\textbf{Average Rows per Table}} & \multicolumn{4}{|c}{\textbf{Average Token per Database}} \\
 \hline
 \textbf{Database Name} & \textbf{8K} & \textbf{16K} & \textbf{32K} & \textbf{64K} & \textbf{8K} & \textbf{16K} & \textbf{32K} & \textbf{64K} \\
 \hline
airline & 28.60 & 48.30 & 80.07 & 134.17 & $5.07\times 10^3$ & $9.45\times 10^3$ & $1.78\times 10^4$ & $3.42\times 10^4$ \\
food\_inspection & 31.97 & 63.63 & 126.33 & 250.50 & $5.87\times 10^3$ & $1.16\times 10^4$ & $2.27\times 10^4$ & $4.46\times 10^4$ \\
movie & 23.83 & 47.17 & 92.60 & 180.40 & $5.62\times 10^3$ & $1.10\times 10^4$ & $2.11\times 10^4$ & $4.09\times 10^4$ \\
music\_tracker & 95.90 & 191.95 & 382.70 & 765.25 & $4.92\times 10^3$ & $9.70\times 10^3$ & $1.95\times 10^4$ & $3.89\times 10^4$ \\
restaurant & 79.50 & 149.40 & 284.07 & 718.93 & $5.16\times 10^3$ & $9.82\times 10^3$ & $1.89\times 10^4$ & $4.85\times 10^4$ \\
university & 47.28 & 83.07 & 145.32 & 353.55 & $5.33\times 10^3$ & $9.60\times 10^3$ & $1.74\times 10^4$ & $4.49\times 10^4$ \\
cookbook & 15.82 & 29.82 & 60.62 & 114.28 & $5.88\times 10^3$ & $1.10\times 10^4$ & $2.27\times 10^4$ & $4.28\times 10^4$ \\
food\_facility\_inspections & 24.00 & 47.97 & 95.80 & 190.70 & $5.45\times 10^3$ & $1.06\times 10^4$ & $2.10\times 10^4$ & $4.13\times 10^4$ \\
water\_quality & 17.80 & 35.67 & 70.50 & 137.93 & $5.31\times 10^3$ & $1.04\times 10^4$ & $2.02\times 10^4$ & $3.92\times 10^4$ \\
global\_biodiversity & 32.00 & 64.00 & 128.00 & 256.00 & $5.40\times 10^3$ & $1.06\times 10^4$ & $2.09\times 10^4$ & $4.17\times 10^4$ \\
\hline
Overall & 39.67 & 76.10 & 146.60 & 310.17 & $5.40\times 10^3$ & $1.04\times 10^4$ & $2.02\times 10^4$ & $4.17\times 10^4$ \\
\hline
\end{tabular}%
\label{tab:sampled_table_info}

\end{table*}

\subsection{Evaluation Task Categories}
\label{sec:task_cate}

The landscape of table QA benchmarks has evolved substantially over time, reflecting increasingly sophisticated LLM capabilities. Early benchmarks emphasized relatively straightforward tasks, primarily involving direct table lookups and aggregations, which required extracting values or computing basic summaries from tabular data~\cite{wikitablequestion,openwikitable}. As research advanced, benchmarks began to incorporate more intricate tasks demanding numerical reasoning, such as arithmetic operations and understanding numerical relationships, thereby elevating task complexity and sophistication~\cite{finqa,multihiertt}.

Despite these advancements, most current datasets are limited to short-context, single-table scenarios, focusing heavily on analysis within constrained contexts. While they frequently include multi-step arithmetic tasks like addition, subtraction, multiplication, or division~\cite{finqa,multihiertt}, they rarely capture the complexities inherent in long-context, multi-table situations. To address this limitation, our benchmark is explicitly structured around three carefully defined categories—\textit{lookup}, \textit{aggregation}, and \textit{complex calculation}—corresponding to distinct levels of difficulty. This categorization enables a comprehensive assessment across various table QA complexities and scenarios.

\begin{figure*}[t!]
\centering

\includegraphics[width=0.85\textwidth]{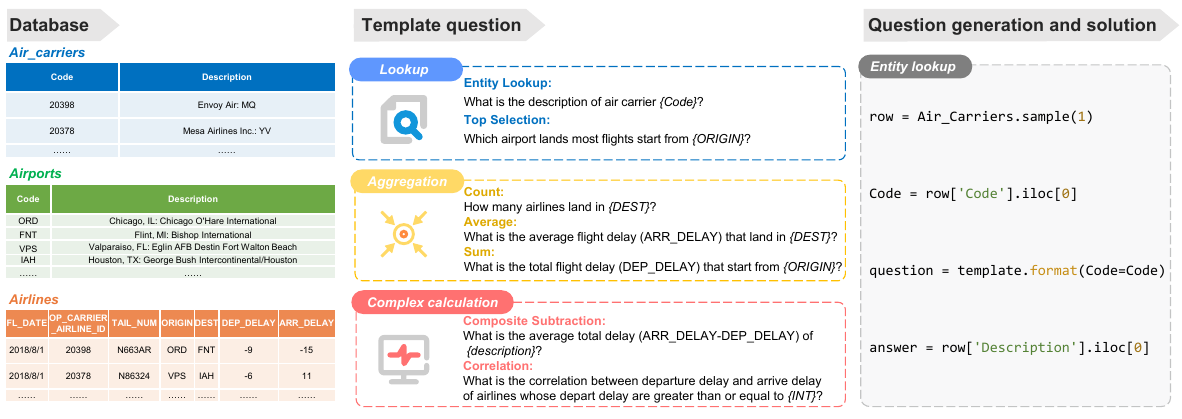}

\caption{Benchmark construction example: a visual illustration of symbolic extension and question-category formation in the ``airline'' database.}
\label{fig:formation}
\vspace{-1.25em}
\end{figure*}

{Let the database be given as $\mathcal{D=(R, E)}$ . We use standard relational algebra $\sigma, \pi, \bowtie, \gamma$ for selection, projection, natural join, and group-by aggregation; $\text{COUNT, SUM}$ for standard aggregates; and $\text{COR}$ for the correlation between 2 numeric selected columns. Given a condition $\Theta$ and a set of attributes $A$, let $\bowtie (A,\Theta)$ denote the minimal join closure that contains all attributes required by $\Theta$ and the output. The formal definitions of each subcategory are as follows:}

\begin{itemize} [topsep=5pt, leftmargin=1em]
    \item \textit{Lookup} tasks are foundational in table-based reasoning. They require the model to locate and extract specific information from tables. We design two tasks in this category:
    
    \begin{itemize}[topsep=0pt, leftmargin=1em]
        \item[$\circ$] \textit{Entity lookup} task retrieves a single specific value in the table directly based on given conditions. {Given target attribute $a$ and condition $\Theta$,

        $$\text{EL}(a,\Theta)=\pi_a(\sigma_\Theta(\bowtie(\{a\},\Theta)))$$

        Condition $\Theta$ ensures the answer is a single item.}

        \item[$\circ$] \textit{Top selection} task focuses on identifying key elements or the top entities in a table based on a specific criterion. {Given grouping key $g$, a metric $m=\text{AGG}(e)$ (e.g., $\text{COUNT, SUM}$) and condition $\Theta$,
        
        $$G=\gamma_{g;c:=\text{AGG}(e)}(\sigma_\Theta(\bowtie(\{g,e\},\Theta)))$$
        $$\text{TS}(g,\Theta,\text{AGG}(e))=\pi_g(\sigma_{c=\max\pi_c}(G))$$
        }
    \end{itemize}

    \item \textit{Aggregation} tasks, though conceptually simpler, test an LLM’s ability to filter and compute integrated information from the table or the join of multiple tables. We include three aggregation functions in categories:
    
    \begin{itemize}[topsep=0pt, leftmargin=1em]
        \item[$\circ$] \textit{Count} task requires the model to determine the total number of rows or elements satisfying a specific condition. {For a specific condition $\Theta$,

        $$\text{CNT}(\Theta)=\text{COUNT}(\sigma_\Theta(\bowtie(\varnothing,\Theta)))$$
        }

        \item[$\circ$] \textit{Sum} task asks the LLM to compute the sum of a specific numerical attribute across the rows that meet certain criteria. {For a numerical column $a$ and condition $\Theta$,

        $$\text{SUM}(a,\Theta)=\text{SUM}(\pi_a(\sigma_\Theta(\bowtie(\{a\},\Theta))))$$
        }

        \item[$\circ$] \textit{Average} task requires the LLM to calculate the mean of a numerical column for rows matching conditions. {For a numerical column $a$ and condition $\Theta$,

        $$\text{AVG}(a,\Theta)=\frac{\text{SUM}(a,\Theta)}{\text{CNT}(\Theta)}$$
        }
        
    \end{itemize}

    \item \textit{Complex calculation} tasks evaluate advanced reasoning capabilities, focusing on more intricate operations. We categorize these into two subcategories:  
    \begin{itemize}[topsep=0pt, leftmargin=1em]
        \item[$\circ$] \textit{Composite comparison} task requires the LLM to compare the difference between two values, which may either be directly available in the table or derived through intermediate calculations. {For a comparison expression $e$ (e.g., $e=\text{ARR\_DELAY-DEP\_DELAY}$), a metric $m=\text{AGG}(e)$ and condition $\Theta$,

        $$\text{CC}(\text{AGG(e)},\Theta)=\text{AGG}(\pi_a(\sigma_\Theta(\bowtie(\{e\},\Theta))))$$
        }
        
        \item[$\circ$] \textit{Correlation} task requires the LLM to compute the statistics between two numeric columns. For numeric columns $a$, $b$ and condition $\Theta$, the answer is the Pearson
correlation coefficient over the rows satisfying $\Theta$:
\[
\mathrm{COR}(a,b,\Theta)
= \frac{\sum_{i \in I_\Theta} (a_i - \bar{a})(b_i - \bar{b})}{
\sqrt{\sum_{i \in I_\Theta} (a_i - \bar{a})^2}
\sqrt{\sum_{i \in I_\Theta} (b_i - \bar{b})^2}},
\]
where $I_\Theta$ indexes rows satisfying $\Theta$, and $\bar{a}$, $\bar{b}$ are the corresponding
sample means.

    \end{itemize}
\end{itemize}

Unlike existing Table QA benchmarks, our tasks are clearly categorized across multiple tables, enabling targeted question creation and systematic performance evaluation. By organizing tasks hierarchically - from simple lookups to complex calculations - our benchmark compares model performance across a spectrum of challenges. It also emphasizes scalability and multi-table contexts, filling critical gaps in current datasets while maintaining practical reasoning depth. This structure enhances evaluation robustness and promotes the model's capabilities of handling intricate situations.

\vspace{-0.5em}
\subsection{Question Generation by Symbolic Extension}
\label{sec:symbolic}

Inspired by the GSM-Symbolic framework~\cite{mirzadeh2024gsm}, we adopt symbolic extension in our benchmark to generate high-quality evaluation questions. By combining symbolic extension with sampling, we create diverse and meaningful queries, enhancing the benchmark's robustness for LLM evaluation.

\noindent\textbf{Symbolic Question Generation.}
Our symbolic extension is divided into two principal components: template question design, and the generation of questions and solutions, as depicted in Fig.~\ref{fig:formation}. Template questions are crafted with placeholder variables instead of fixed values, enabling dynamic content generation. These variables are subsequently instantiated, and the correct answers are computed using Python implementation. This methodology facilitates the creation of multiple question instances from a single template, thereby enhancing the benchmark's versatility and scalability.
To enable a more effective evaluation, we employ multiple-choice questions (MCQs) rather than relying solely on traditional metrics such as exact match, BLEU, or F1 scores. These conventional metrics can fall short of accurately assessing the reasoning capabilities of LLMs, while MCQs provide discrete, comparable options that support scalable and unambiguous scoring~\cite{balepur2024your}. At the same time, prior work has also noted that option-based evaluation can mask part of the difficulty of open-ended answer generation~\cite{zheng2023large}. We therefore keep MCQ as the primary benchmark format and later report a small aligned open-ended pilot as a supplementary check.
For each database, we manually design two template questions per subcategory, each paired with a corresponding Python code solution. Using two templates keeps each reasoning type expressed through more than one question formulation, while still keeping manual template design and answer verification tractable. This approach yields a total of $140$ template questions across all databases and subcategories. To populate the benchmark with diverse instances, we leverage the ten database instances created for each database and context length. Using the symbolic extension, we generate ten question instances for each template question. For each context length, this yields $10$ databases $\times$ $10$ sampled database instances $\times$ $7$ question subcategories $\times$ $2$ templates $\times$ $10$ symbolic instantiations $= 14{,}000$ question instances. An overview of the total number of benchmark instances is provided in TABLE~\ref{tab:count}, illustrating the extensive scale and scope of our benchmark.
More generation examples for the Airline database are provided in Appendix~\S{IV}.

\begin{table}[htbp]
\centering

\caption{An overview of our benchmark instances.}

\resizebox{\columnwidth}{!}{
\begin{tabular}{l c c} 
 \hline
 \textbf{Context Length} & \textbf{Database Instances} & \textbf{Question Instances} \\
 \hline
 8K & 100 & 14000 \\
 16K & 100 & 14000 \\
 32K & 100 & 14000 \\
 64K & 100 & 14000 \\
 \hline
 Total & 400 & 56000 \\
 \hline
\end{tabular}
}
\label{tab:count}
\vspace{-1em}
\end{table}

\noindent \textbf{Wrong Choice Generation}.
To create incorrect options for the MCQs, we use a rule-based approach. For \textit{entity lookup} and \textit{top selection} tasks, we randomly select different cells from the same column to generate error choices. For the rest tasks, which require numerical answers, we produce three error options by multiplying the correct answer by $0.25$, $2.0$, and $3.0$. While this method is simple, our experiments show that these questions remain challenging for LLMs, especially due to the design consideration that our tasks require reasoning over long contexts and multiple tables.

\section{Evaluation Setup}

Our benchmark is created using database sampling and symbolic extension, making it challenging to test all the generated instances. To address this, we design experiments across multiple dimensions to evaluate its performance effectively.

\subsection{LLM Benchmark Scope}

To ensure a comprehensive evaluation of LLM's performance on our benchmark, we select 30 LLMs from various companies or research organizations. The selected models cover the most advanced proprietary LLMs available such as \textsc{GPT}~\cite{gpt4o2024,gpto3mini,gpto1mini} and \textsc{Gemini}~\cite{team2023gemini}, as well as other widely-recognized open-source models such as \textsc{Qwen2.5}~\cite{qwen2.5}. It is worth mentioning that we also choose a strong model from DeepSeek that employs mixture-of-experts (MoE) architecture{~\cite{deepseekai2024deepseekv3technicalreport,guo2025deepseek}}. Moreover, we include two domain-specific LLMs, \textsc{TableLlama}{~\cite{zhang2024tablellama}} and \textsc{TableGPT2}{~\cite{su2024tablegpt2}}, which are specifically fine-tuned for analyzing tabular data and accomplishing various table-based tasks. Meanwhile, the parameter scales of the models we choose range from 2B to 671B, which may provide insights into the relationship between model size and multi-table QA performance. Such diversity ensures the benchmark evaluates models of varying architectures, specializations, and computational complexities, providing valuable insights into the strengths and limitations of current LLMs. The details of the selected LLMs are in the Appendix~\S{I}.

\subsection{Evaluation Design and Implementation}
\label{sec:eval_design}

We configure our evaluation setup to be simple and consistent, ensuring it provides consistent evaluation results.

\noindent\textbf{Design of the Experiment}. We design a series of concrete experiments to answer the following questions:

\begin{itemize}[topsep=5pt, leftmargin=1em]

\item (\underline{i}) \textit{How do single- and multi-table settings differ?}

\item (\underline{ii}) \textit{How does the choice of table serialization format impact LLM performance in multi-table question answering tasks?}

\item (\underline{iii}) \textit{How do different categories of LLMs vary in their ability in multi-table QA under different context lengths?}

\item 
(\underline{iv}) \textit{How does the symbolic extension with sampling improve the diversity and difficulty of generated questions?}

\item 
(\underline{v}) \textit{What is the performance variance between direct LLM prompt and LLM-based Text2SQL approaches?}
\end{itemize}

Concretely, each experiment aims to evaluate one particular facet of the above questions, as we enumerated below:

\begin{itemize}[topsep=5pt, leftmargin=1em]
\item\textbf{Experiment 1. Comparison of single- and multi-table scenarios}: previous table QA datasets predominantly focus on single-table settings. To investigate how integrating multiple tables affects question-answering performance, this experiment compares the performance of LLMs when using single versus multiple tables. Specifically, the LLMs tested include \textsc{GLM-4-9B-Chat}, \textsc{Qwen2.5-7B-Instruct}, \textsc{Llama3.1-8B-Instruct}, \textsc{GPT-4o-mini} and \textsc{
DeepSeek-R1-Distill-Qwen-7B}. In the single-table condition, all relevant tables are merged into one table before being provided to the LLMs.

\item \textbf{Experiment 2. Serialization format evaluation}: given the importance of serialization in managing long-context, multi-table data, our experiment compares four commonly used formats: Markdown, CSV, JSON and HTML. These formats are selected for their standardization. We conduct this comparison using a variety of models including \textsc{Qwen2.5-7B-Instruct}, \textsc{Qwen2.5-Coder-7B-Instruct}, and \textsc{Llama3.1-8B-Instruct}. The results will determine the serialization method in subsequent experiments to ensure consistency across all evaluations.

\item \textbf{Experiment 3. Comprehensive evaluations of LLMs}: following the selection of the serialization method, we undertake a comprehensive evaluation of LLMs from various providers and of different scales. This assessment includes several open source LLMs including \textsc{Qwen2.5}, \textsc{Llama3.1}, \textsc{Baichuan2}, \textsc{GLM-4}, \textsc{Mistral}, \textsc{DeepSeek-V3}, \textsc{DeepSeek-R1}, and \textsc{TableGPT2}, tested at four context lengths of 8K, 16K, 32K, and 64K. Other open source LLMs, including \textsc{Gemma2}, \textsc{TableLlama}, and \textsc{Vicuna}, are tested up to their respective context length limits. Close-source models are included like \textsc{GPT-4o}, \textsc{GPT-4o-mini}, \textsc{GPT-o1-mini}, \textsc{GPT-o3-mini}, \textsc{GPT-5.1}, and \textsc{Gemini-3.1-Pro-Preview}. We also evaluate \textsc{GPT-5.1} under two settings: Direct Prompting (without tools) and an Agentic setting (with tool calls). In the Agentic setting, we allow \textsc{GPT-5.1} to autonomously invoke a code interpreter tool.
This experiment allows us to observe performance trends and compare capabilities under uniform conditions across different LLMs.

\item\textbf{Experiment 4. Influence of sampling and symbolic extension}: this experiment focuses on assessing how well sampling and symbolic extensions contribute to enhancing the complexity of the benchmark. This targeted evaluation helps us understand the effectiveness of sampling and symbolic extensions in enriching the benchmark's capacity to challenge and measure the analytic abilities of LLMs.

\item\textbf{Experiment 5. Compare direct LLM prompts with LLM-based Text2SQL}: Given the prevalence of relational databases in real-world applications, LLM-based Text2SQL becomes a popular paradigm in production. We attempt to understand whether directly prompting LLMs or utilizing LLM-based Text2SQL approaches yields superior performance in multi-table QA tasks. This experiment systematically evaluates both methods, highlighting their respective strengths and limitations, particularly in handling complex queries and varying context lengths.

\end{itemize}

This step-by-step evaluation ensures that our benchmark is rigorous, consistent, and provides meaningful insights into the performance of different LLMs in table QA tasks. We attach additional evaluation details in Appendix~\S{II}.

\noindent \textbf{Accuracy Metric}. For Experiments 1--4, accuracy is the percentage of questions answered correctly. In the primary MCQ setting, a prediction is counted as correct only when the selected option matches the gold option for that question. Per-question-type columns report the accuracy within each subcategory, and the \textit{Overall} column reports the aggregate score. Because our sampling is balanced across the seven question types, this overall score is numerically identical to the mean over the per-question-type accuracies.

\noindent \textbf{Text2SQL Evaluation Protocol}. In Experiment 5, each model first generates SQL from the database schema and the natural-language question. We then execute the generated SQL on the corresponding sampled database instance and compare the returned final answer with the benchmark label. We evaluate final answers rather than SQL strings, because many TQA-Bench items require numerical analytical outputs and different SQL formulations can still lead to equivalent answers. We use a task-aware answer-layer protocol for this comparison, and the detailed matching rules are given in Appendix~\S{II}.

\noindent \textbf{Supplementary Open-Ended Pilot}. As a complementary check beyond MCQ, we additionally evaluate a small aligned open-ended value-generation setting on \textsc{GPT-4o-mini} at 8K. This pilot uses the same 700 questions as the aligned MCQ subset, removes the answer options, and asks the model to produce a free-form response ending with \texttt{"Final answer: ..."}. The extracted final answer is then compared with the benchmark label using a task-aware answer-layer protocol described in Appendix~\S{II}.

\noindent \textbf{Statistics of the Benchmark Tasks}. To manage computational costs, we determine the number of test cases as below:
\textbf{Experiments 1 and 4} are conducted on the ``airline'' database at the 8K context length, utilizing all ten sampled database instances and $1400$ generated questions.
In \textbf{Experiments 2 3 and 5}, we test five database instances for each context length and database. For each database instance, one question instance is selected for each of the 14 question templates. This results in 50 database instances and 700 questions for each context length. This strategy keeps the testing practical while still providing representative results.

\section{Results and Analysis}

We enumerate the concrete experimental results to answer the question proposed in Section \S\ref{sec:eval_design}. Additional evaluation details, paired significance tests, and contrastive case analyses are provided in Appendix~\S{II} and Appendix~\S{III}.

\vspace{-1em}
\subsection{Compare Single- and Multi- Table Scenarios}

We first compared LLM performance on single-table versus multi-table scenarios to assess how table structure affects accuracy and to validate the need of multi-table evaluations. For a controlled comparison, we merged the necessary tables into a single table for the single-table scenario, while retaining the original separate tables in the multi-table setting.

\noindent \textbf{Comparison Results and Discussion.}
TABLE~\ref{tab:single-multi-compare} shows that LLMs perform better in the single-table scenario, with up to $20\%$ higher accuracy, which indicates that single-table benchmark lacks the ability to evaluate LLMs under complex analytic tasks. This finding confirms the demand for multi-table benchmarks.
Our results also suggest that converting multi-table queries into single-table ones through merging can boost performance if done carefully. Merges must be semantically meaningful (e.g., renaming ``\texttt{Description}'' to ``\texttt{ORIGIN\_Description}'' and ``\texttt{DEST\_Description}'' when merging ``\texttt{Airlines}'' with ``\texttt{Airports}'') and limited to relevant tables to avoid excessive context length.

\begin{table*}[htbp]
    \centering
    \caption{Single-Table and multi-tables accuracy comparison.}
    
    \begin{tabular}{cccccc}
        \hline
        \textbf{Model}    & \textbf{GLM-4-9B-Chat} & \textbf{Qwen-2.5-7B-Instruct} & \textbf{Llama3.1-8B-Instruct} & \textbf{GPT-4o-mini} & \textbf{DS-R1-Qwen-7B} \\
        \hline
        single-table & 32.43 & 55.36 & 49.21 & 54.50 & 34.00 \\
        multi-tables & 22.14 & 35.29 & 31.79 & 48.43 & 28.79 \\
        \hline
    \end{tabular}
    \label{tab:single-multi-compare}
\end{table*}

A paired McNemar analysis on aligned cases from the \texttt{airline} database at 8K further confirms that this gap is systematic rather than incidental. Across the primary models used for the mechanism analysis, the single-table advantage is statistically significant (all $p<0.005$). The main penalties are concentrated on \textit{Count} and \textit{Top Selection}; for weaker models, the degradation further extends to \textit{Sum}, \textit{Average}, and \textit{Composite Comparison}, while \textit{Correlation} does not exhibit a stable single-vs.-multi penalty. Appendix~\S{III} reports the full paired statistics and a focused \textsc{GPT-4o-mini} case analysis, showing that the main bottleneck lies in cross-table entity grounding, join scope identification, and aggregation over the intended result set.

\begin{tcolorbox}[colback=blue!5!white,colframe=blue!75!black]
  \textbf{Answer to Question (\underline{i})}:
\textit{Our findings show that LLMs generally achieve better performance in the single-table scenario, and paired significance tests confirm that this is a systematic effect rather than a random result. The single-vs.-multi penalty is concentrated on tasks that require cross-table grounding before aggregation or ranking, demonstrating that single-table benchmarks alone are insufficient to evaluate complex real-world Table QA applications. On the other hand, the transformation of multi-table input to single-table presents a potential avenue for improving model performance in multi-table scenarios, if merging strategies can preserve semantic integrity.}
\end{tcolorbox}
\vspace{-1em}

\subsection{Serialization Format Evaluation Results}

We compare four serialization formats—Markdown, CSV, JSON, and HTML—to assess their impact on LLM performance. Before running the experiments, we measure the tokenized context length of each format, where all formats are serialized consistently using \texttt{pandas}. TABLE~\ref{tab:format_count} reports the detailed results. Token costs vary widely across formats even for the same database: CSV is the most token-efficient format, while HTML requires nearly three times more tokens. Among them, JSON and HTML represent mainstream semi-structured encodings, which typically introduce additional structural tokens compared to the more lightweight CSV/Markdown formats. In some cases, a 64K-scale database serialized in HTML exceeds 128K tokens, surpassing most typical LLM limits; thus, we evaluate HTML only up to maximum 32K. TABLE~\ref{tab:serialize} shows model performance across formats.

\begin{table}[htbp]
    \centering
    \caption{Context length of different formats and scales.}
    
    \begin{tabular}{ccccc}
        \hline
        Format & 8K & 16K & 32K & 64K\\
        \hline
        Markdown & $5.40\times 10^3$ & $1.04\times 10^4$ & $2.02\times 10^4$ & $4.17\times 10^4$\\
        CSV & $3.73\times 10^3$ & $7.24\times 10^3$ & $1.42\times 10^4$ & $2.93\times 10^4$\\
        JSON & $5.75\times 10^3$ & $1.12\times 10^4$ & $2.19\times 10^4$ & $4.54\times 10^4$\\
        HTML & $1.05\times 10^4$ & $2.02\times 10^4$ & $3.92\times 10^4$ & $8.16\times 10^4$\\
        \hline
    \end{tabular}
    \label{tab:format_count}
\end{table}

\begin{table*}[htbp]
\centering
\caption{Results of the multi-table serialization format comparison. Accuracies are computed at the question level.} 
\resizebox{\textwidth}{!}{%
\begin{tabular}{ ll|ccccccc|c|ccccccc|c  }
 \hline
 & & \multicolumn{8}{|c|}{\textbf{8K}} & \multicolumn{8}{|c}{\textbf{16K}} \\
 \hline
 \textbf{Model} & \textbf{Format} & \textbf{EL} & \textbf{TS} & \textbf{CNT} & \textbf{SUM} & \textbf{AVG} & \textbf{CC} & \textbf{COR} & \textbf{Overall} & \textbf{EL} & \textbf{TS} & \textbf{CNT} & \textbf{SUM} & \textbf{AVG} & \textbf{CC} & \textbf{COR} & \textbf{Overall} \\
\hline
\multirow{4}{12em}{Qwen2.5-7B-Instruct} & MD & 86.00 & 52.00 & 42.00 & 45.00 & 50.00 & 59.00 & 15.00 & 49.86 & 84.00 & 53.00 & 54.00 & 36.00 & 52.00 & 58.00 & 19.00 & 50.86 \\
 & CSV & 75.00 & 36.00 & 25.00 & 38.00 & 38.00 & 61.00 & 11.00 & 40.57 & 71.00 & 41.00 & 30.00 & 27.00 & 24.00 & 51.00 & 14.00 & 36.86 \\
 & JSON & 61.00 & 42.00 & 38.00 & 33.00 & 42.00 & 52.00 & 19.00 & 41.00 & 63.00 & 46.00 & 29.00 & 17.00 & 25.00 & 42.00 & 14.00 & 33.71 \\
 & HTML & 82.00 & 43.00 & 40.00 & 33.00 & 41.00 & 66.00 & 21.00 & 46.57 & 78.00 & 55.00 & 29.00 & 24.00 & 26.00 & 52.00 & 31.00 & 42.14 \\
\hline
\multirow{4}{12em}{Qwen2.5-Coder-7B-Instruct} & MD & 87.00 & 51.00 & 40.00 & 26.00 & 44.00 & 62.00 & 26.00 & 48.00 & 83.00 & 55.00 & 34.00 & 31.00 & 40.00 & 63.00 & 24.00 & 47.14 \\
 & CSV & 88.00 & 50.00 & 43.00 & 32.00 & 42.00 & 62.00 & 16.00 & 47.57 & 80.00 & 43.00 & 30.00 & 28.00 & 38.00 & 59.00 & 23.00 & 43.00 \\
 & JSON & 59.00 & 35.00 & 34.00 & 22.00 & 31.00 & 55.00 & 20.00 & 36.57 & 64.00 & 40.00 & 33.00 & 18.00 & 30.00 & 49.00 & 24.00 & 36.86 \\
 & HTML & 81.00 & 45.00 & 38.00 & 27.00 & 39.00 & 61.00 & 23.00 & 44.86 & 81.00 & 47.00 & 31.00 & 24.00 & 39.00 & 54.00 & 26.00 & 43.14 \\
\hline
\multirow{4}{12em}{Llama3.1-8B-Instruct} & MD & 86.00 & 62.00 & 32.00 & 30.00 & 44.00 & 60.00 & 12.00 & 46.57 & 84.00 & 58.00 & 29.00 & 20.00 & 34.00 & 56.00 & 20.00 & 43.00 \\
 & CSV & 87.00 & 55.00 & 32.00 & 22.00 & 33.00 & 66.00 & 15.00 & 44.29 & 81.00 & 55.00 & 28.00 & 16.00 & 21.00 & 57.00 & 13.00 & 38.71 \\
 & JSON & 65.00 & 37.00 & 26.00 & 19.00 & 24.00 & 49.00 & 20.00 & 34.29 & 70.00 & 53.00 & 19.00 & 12.00 & 17.00 & 45.00 & 10.00 & 32.29 \\
 & HTML & 82.00 & 48.00 & 34.00 & 25.00 & 31.00 & 65.00 & 23.00 & 44.00 & 80.00 & 56.00 & 20.00 & 21.00 & 19.00 & 56.00 & 18.00 & 38.57 \\
 \hline
 & & \multicolumn{8}{|c|}{\textbf{32K}} & \multicolumn{8}{|c}{\textbf{64K}} \\
 \hline
\multirow{4}{12em}{Qwen2.5-7B-Instruct} & MD & 67.00 & 51.00 & 32.00 & 28.00 & 33.00 & 45.00 & 35.00 & 41.57 & 44.00 & 49.00 & 26.00 & 22.00 & 26.00 & 32.00 & 35.00 & 33.43 \\
 & CSV & 66.00 & 41.00 & 15.00 & 10.00 & 15.00 & 52.00 & 37.00 & 33.71 & 58.00 & 38.00 & 23.00 & 18.00 & 12.00 & 42.00 & 37.00 & 32.57 \\
 & JSON & 44.00 & 39.00 & 21.00 & 14.00 & 19.00 & 27.00 & 26.00 & 27.14 & 41.77 & 62.03 & 26.58 & 13.92 & 15.19 & 28.21 & 21.79 & 29.95 \\
 & HTML & 62.00 & 41.00 & 28.00 & 20.00 & 24.00 & 39.00 & 38.00 & 36.00 & OOC & OOC & OOC & OOC & OOC & OOC & OOC & OOC \\
\hline
\multirow{4}{12em}{Qwen2.5-Coder-7B-Instruct} & MD & 75.00 & 51.00 & 33.00 & 24.00 & 38.00 & 51.00 & 42.00 & 44.86 & 63.00 & 50.00 & 20.00 & 19.00 & 31.00 & 40.00 & 35.00 & 36.86 \\
 & CSV & 76.00 & 40.00 & 17.00 & 20.00 & 23.00 & 38.00 & 27.00 & 34.43 & 60.00 & 43.00 & 26.00 & 19.00 & 32.00 & 38.00 & 34.00 & 36.00 \\
 & JSON & 49.00 & 43.00 & 30.00 & 20.00 & 20.00 & 39.00 & 33.00 & 33.43 & 45.00 & 43.00 & 23.00 & 12.00 & 25.00 & 20.00 & 27.00 & 27.86 \\
 & HTML & 66.00 & 46.00 & 22.00 & 24.00 & 39.00 & 47.00 & 34.00 & 39.71 & OOC & OOC & OOC & OOC & OOC & OOC & OOC & OOC \\
\hline
\multirow{4}{12em}{Llama3.1-8B-Instruct} & MD & 78.00 & 49.00 & 20.00 & 12.00 & 17.00 & 50.00 & 22.00 & 35.43 & 74.00 & 54.00 & 21.00 & 10.00 & 18.00 & 45.00 & 29.00 & 35.86 \\
 & CSV & 73.00 & 50.00 & 19.00 & 15.00 & 9.00 & 44.00 & 21.00 & 33.00 & 75.00 & 55.00 & 21.00 & 17.00 & 11.00 & 41.00 & 30.00 & 35.71 \\
 & JSON & 51.00 & 42.00 & 14.00 & 12.00 & 9.00 & 37.00 & 14.00 & 25.57 & 57.47 & 45.98 & 16.28 & 13.95 & 10.47 & 29.07 & 19.77 & 27.65 \\
 & HTML & 75.00 & 48.00 & 15.00 & 13.00 & 18.00 & 54.00 & 32.00 & 36.43 & OOC & OOC & OOC & OOC & OOC & OOC & OOC & OOC \\
 \hline
\end{tabular}%
}
\label{tab:serialize}
\end{table*}

\noindent\textbf{Overall Results.}
The benchmark results indicate that \textit{Markdown consistently yields better performance than other formats for most LLMs}. This advantage holds across the majority of context lengths, suggesting that Markdown offers a more readable and structured representation for table analysis. By contrast, the relative ordering between CSV and HTML is less stable and varies by model and scale, making it hard to declare a clear winner. Finally, JSON performs the worst: it attains the lowest accuracy for almost all models and context lengths, indicating that its representation is generally less suitable for multi-table QA in our setting.

\noindent\textbf{Detailed Discussion.} We further enumerate several interesting observations. \underline{First}, \textit{when analyzing results across various subcategories, the advantages of Markdown remain consistently}. Although some LLMs exhibited improved performance with alternative formats in certain subcategories and context lengths, the performance differences were generally minor. In contrast, Markdown consistently provided a more substantial accuracy boost for most LLMs, reinforcing its advantage as a serialization format. \underline{Second}, \textit{our analysis revealed that coder LLMs outperformed their original counterparts with certain formats}. Notably, in the CSV format, the \textsc{Qwen2.5-Coder-7B-Instruct} outperformed \textsc{Qwen2.5-7B-Instruct} across all scales. For other formats, performance improvements were observed in specific scales. Note that the other experiments use Markdown, ensuring optimal performance.

\begin{tcolorbox}[colback=blue!5!white,colframe=blue!75!black]
  \textbf{Answer to Question (\underline{ii})}:
\textit{Our findings consistently suggest that Markdown is the superior format for table serialization for LLMs in complex multi-table QA tasks, compared with CSV, JSON, and HTML.}
\end{tcolorbox}
\vspace{-1em}

\subsection{Comprehensive LLM Evaluation Results}

The primary goal of our benchmark is to provide a comprehensive assessment of LLM performance on multi-table QA tasks. In this section, we report the detailed evaluation results of the tested LLMs in TABLE~\ref{tab:broader_test}.

\begin{table*}[!t]
\centering
\caption{Complete multi-table benchmark results. NFI indicates ``not following instructions'', OOC indicates ``out of context''.}

\resizebox{0.95\textwidth}{!}{%
\begin{tabular}{ l|ccccccc|c|ccccccc|c  }
 \hline
 & \multicolumn{8}{c}{\textbf{8K}} & \multicolumn{8}{|c}{\textbf{16K}} \\
 \hline
 \textbf{Model} & \textbf{EL} & \textbf{TS} & \textbf{CNT} & \textbf{SUM} & \textbf{AVG} & \textbf{CC} & \textbf{COR} & \textbf{Overall} & \textbf{EL} & \textbf{TS} & \textbf{CNT} & \textbf{SUM} & \textbf{AVG} & \textbf{CC} & \textbf{COR} & \textbf{Overall} \\
 \hline
 \rowcolor{lightgray}
 \multicolumn{17}{c}{Chat Models} \\
 \hline
GLM-4-9B-Chat & 56.00 & 56.00 & 27.00 & 16.00 & 19.00 & 47.00 & 13.00 & 33.43 & 57.00 & 54.00 & 23.00 & 16.00 & 6.00 & 44.00 & 18.00 & 31.14 \\
Baichuan2-7B-Chat & 35.00 & 34.00 & 22.00 & 19.00 & 20.00 & 17.00 & 22.00 & 24.14 & 26.00 & 45.00 & 14.00 & 11.00 & 14.00 & 17.00 & 16.00 & 20.43 \\
Baichuan2-13B-Chat & 2.00 & 4.00 & 8.00 & 2.00 & 4.00 & 3.00 & 7.00 & 4.29 & 5.00 & 3.00 & 1.00 & 6.00 & 3.00 & 4.00 & 5.00 & 3.86 \\
Vicuna-7B-V1.5-16K & 29.00 & 22.00 & 40.00 & 20.00 & 23.00 & 30.00 & 27.00 & 27.29 & 11.00 & 23.00 & 19.00 & 11.00 & 13.00 & 13.00 & 8.00 & 14.00 \\
Vicuna-13B-V1.5-16K & 31.00 & 29.00 & 12.00 & 2.00 & 8.00 & 10.00 & 12.00 & 14.86 & 19.00 & 26.00 & 19.00 & 3.00 & 7.00 & 9.00 & 4.00 & 12.43 \\
\hline
\rowcolor{lightgray}
\multicolumn{17}{c}{Instruct Models} \\
\hline
Mistral-7B-Instruct & 44.00 & 41.00 & 16.00 & 8.00 & 13.00 & 25.00 & 12.00 & 22.71 & 46.00 & 48.00 & 14.00 & 5.00 & 10.00 & 18.00 & 4.00 & 20.71 \\
Mistral-Nemo-Instruct & 83.00 & 59.00 & 38.00 & 32.00 & 40.00 & 57.00 & 33.00 & 48.86 & 42.00 & 43.00 & 21.00 & 13.00 & 17.00 & 34.00 & 18.00 & 26.86 \\
Llama3.1-8B-Instruct & 86.00 & 62.00 & 32.00 & 30.00 & 44.00 & 60.00 & 12.00 & 46.57 & 84.00 & 58.00 & 29.00 & 20.00 & 34.00 & 56.00 & 20.00 & 43.00 \\
Llama3.1-70B-Instruct & 93.00 & 82.00 & 57.00 & 51.00 & 54.00 & 83.00 & 20.00 & 62.86 & 94.00 & 81.00 & 39.00 & 39.00 & 46.00 & 80.00 & 20.00 & 57.00 \\
Qwen2.5-3B-Instruct & 62.00 & 31.00 & 23.00 & 24.00 & 23.00 & 46.00 & 8.00 & 31.00 & 59.00 & 36.00 & 25.00 & 15.00 & 25.00 & 36.00 & 12.00 & 29.71 \\
Qwen2.5-7B-Instruct & 86.00 & 52.00 & 42.00 & 45.00 & 50.00 & 59.00 & 15.00 & 49.86 & 84.00 & 53.00 & 54.00 & 36.00 & 52.00 & 58.00 & 19.00 & 50.86 \\
Qwen2.5-Coder-7B-Instruct & 87.00 & 51.00 & 40.00 & 26.00 & 44.00 & 62.00 & 26.00 & 48.00 & 83.00 & 55.00 & 34.00 & 31.00 & 40.00 & 63.00 & 24.00 & 47.14 \\
Qwen2.5-14B-Instruct & 89.00 & 72.00 & 60.00 & 47.00 & 59.00 & 80.00 & 9.00 & 59.43 & 86.00 & 75.00 & 48.00 & 35.00 & 43.00 & 72.00 & 13.00 & 53.14 \\
Qwen2.5-72B-Instruct & 91.00 & 72.00 & 55.00 & 59.00 & 51.00 & 78.00 & 3.00 & 58.43 & 86.00 & 61.00 & 34.00 & 34.00 & 38.00 & 73.00 & 1.00 & 46.71 \\
Gemma2-2B-It & 49.00 & 33.00 & 28.00 & 25.00 & 13.00 & 18.00 & 31.00 & 28.14 & OOC & OOC & OOC & OOC & OOC & OOC & OOC & OOC \\
Gemma2-9B-It & 76.00 & 41.00 & 42.42 & 28.72 & 27.96 & 46.32 & 2.11 & 38.17 & OOC & OOC & OOC & OOC & OOC & OOC & OOC & OOC \\
Gemma2-27B-It & 83.00 & 44.00 & 29.00 & 31.00 & 33.00 & 65.00 & 15.00 & 42.86 & OOC & OOC & OOC & OOC & OOC & OOC & OOC & OOC \\
DeepSeek-V3 & 94.00 & 89.00 & 79.00 & 84.85 & 79.00 & 88.00 & 48.00 & 80.26 & 94.00 & 86.87 & 74.00 & 79.80 & 70.00 & 81.82 & 39.39 & 75.14 \\
\hline
\rowcolor{lightgray}
\multicolumn{17}{c}{Table Specific Models} \\
\hline
TableGPT2-7B & 80.00 & 49.00 & 37.00 & 24.00 & 33.00 & 48.00 & 23.00 & 42.00 & 68.00 & 63.00 & 37.00 & 11.00 & 27.00 & 38.00 & 27.00 & 38.71 \\
TableLlama & NFI & NFI & NFI & NFI & NFI & NFI & NFI & NFI & OOC & OOC & OOC & OOC & OOC & OOC & OOC & OOC \\
\hline
\rowcolor{lightgray}
\multicolumn{17}{c}{Close-Source Models} \\
\hline
GPT-4o-mini & 82.00 & 66.00 & 55.00 & 51.00 & 60.00 & 72.00 & 40.00 & 60.86 & 82.00 & 71.00 & 49.00 & 47.00 & 57.00 & 68.00 & 24.00 & 56.86 \\
GPT-4o & 92.00 & 88.00 & 80.00 & 76.00 & 76.00 & 90.00 & 48.48 & 78.68 & 91.00 & 82.00 & 75.00 & 63.00 & 68.00 & 85.00 & 42.00 & 72.29 \\
GPT-o1-mini & 90.00 & 84.00 & 85.00 & 89.90 & 75.76 & 77.78 & 73.74 & 82.33 & 87.00 & 88.00 & 78.00 & 84.00 & 73.00 & 77.00 & 51.00 & 76.86 \\
GPT-o3-mini & 89.00 & 86.00 & 91.92 & 93.00 & 85.00 & 91.00 & 86.00 & 88.84 & 92.00 & 91.00 & 94.00 & 94.00 & 84.00 & 93.00 & 84.00 & 90.29 \\
GPT-5.1 (Direct Prompt) & 94.00 & 91.00 & 95.00 & 91.00 & 87.00 & 95.00 & 57.58 & 87.27 & 94.95 & 97.00 & 91.92 & 85.00 & 79.00 & 92.93 & 48.00 & 84.07 \\
GPT-5.1 (Agent) & 93.94 & 91.00 & 95.00 & 92.93 & 85.86 & 93.00 & 60.61 & 87.50 & 95.00 & 94.00 & 91.92 & 90.00 & 82.00 & 94.00 & 48.00 & 84.98 \\
Gemini-3.1-Pro-Preview & 94.00 & 93.00 & 99.00 & 100.00 & 95.00 & 100.00 & 90.00 & 95.86 & 95.00 & 97.00 & 99.00 & 100.00 & 96.00 & 97.00 & 95.00 & 97.00 \\
 \hline
\rowcolor{lightgray}
\multicolumn{17}{c}{Reasoning Models} \\
\hline
DeepSeek-R1-Distill-Qwen-7B & 39.00 & 40.00 & 39.00 & 15.00 & 20.00 & 21.00 & 21.00 & 27.86 & 29.00 & 31.00 & 19.00 & 14.00 & 16.00 & 28.00 & 25.00 & 23.14 \\
DeepSeek-R1-Distill-Qwen-14B & 89.00 & 59.00 & 52.00 & 48.00 & 44.00 & 71.00 & 9.00 & 53.14 & 83.00 & 59.00 & 37.00 & 28.00 & 24.00 & 58.00 & 13.00 & 43.14 \\
DeepSeek-R1 & 94.00 & 93.94 & 95.96 & 98.99 & 94.95 & 98.00 & 84.69 & 94.38 & 94.95 & 97.00 & 93.94 & 98.00 & 94.00 & 97.00 & 69.39 & 92.10 \\
QwQ-32B-Preview & 1.00 & 4.00 & 2.00 & 2.00 & 1.00 & 2.00 & 2.00 & 2.00 & 1.00 & 3.00 & 1.00 & 0.00 & 2.00 & 4.00 & 3.00 & 2.00 \\
 \hline
 & \multicolumn{8}{c}{\textbf{32K}} & \multicolumn{8}{|c}{\textbf{64K}} \\
 \hline
 \rowcolor{lightgray}
 \multicolumn{17}{c}{Chat Models} \\
\hline
GLM-4-9B-Chat & 63.00 & 47.00 & 16.00 & 8.00 & 5.00 & 34.00 & 19.00 & 27.43 & 46.00 & 42.00 & 16.00 & 13.00 & 4.00 & 30.00 & 22.00 & 24.71 \\
Baichuan2-7B-Chat & 27.00 & 33.00 & 12.00 & 13.00 & 14.00 & 13.00 & 12.00 & 17.71 & 28.00 & 41.00 & 21.00 & 16.00 & 22.00 & 14.00 & 17.00 & 22.71 \\
Baichuan2-13B-Chat & 4.00 & 4.00 & 4.00 & 4.00 & 1.00 & 3.00 & 1.00 & 3.00 & 5.00 & 5.00 & 5.00 & 5.00 & 1.00 & 3.00 & 1.00 & 3.57 \\
Vicuna-7B-V1.5-16K & OOC & OOC & OOC & OOC & OOC & OOC & OOC & OOC & OOC & OOC & OOC & OOC & OOC & OOC & OOC & OOC \\
Vicuna-13B-V1.5-16K & OOC & OOC & OOC & OOC & OOC & OOC & OOC & OOC & OOC & OOC & OOC & OOC & OOC & OOC & OOC & OOC \\
\hline
\rowcolor{lightgray}
\multicolumn{17}{c}{Instruct Models} \\
\hline
Mistral-7B-Instruct & 43.00 & 35.00 & 10.00 & 7.00 & 5.00 & 9.00 & 9.00 & 16.86 & OOC & OOC & OOC & OOC & OOC & OOC & OOC & OOC \\
Mistral-Nemo-Instruct & 19.00 & 22.00 & 6.00 & 11.00 & 11.00 & 15.00 & 14.00 & 14.00 & 6.00 & 12.00 & 6.00 & 1.00 & 7.00 & 5.00 & 6.00 & 6.14 \\
Llama3.1-8B-Instruct & 78.00 & 49.00 & 20.00 & 12.00 & 17.00 & 50.00 & 22.00 & 35.43 & 74.00 & 54.00 & 21.00 & 10.00 & 18.00 & 45.00 & 29.00 & 35.86 \\
Llama3.1-70B-Instruct & 93.00 & 73.00 & 29.00 & 24.00 & 33.00 & 76.00 & 27.00 & 50.71 & 88.00 & 83.00 & 26.00 & 25.00 & 27.00 & 57.00 & 29.00 & 47.86 \\
Qwen2.5-3B-Instruct & 46.00 & 28.00 & 19.00 & 11.00 & 21.00 & 22.00 & 15.00 & 23.14 & 39.00 & 33.00 & 11.00 & 10.00 & 19.00 & 24.00 & 19.00 & 22.14 \\
Qwen2.5-7B-Instruct & 67.00 & 51.00 & 32.00 & 28.00 & 33.00 & 45.00 & 35.00 & 41.57 & 44.00 & 49.00 & 26.00 & 22.00 & 26.00 & 32.00 & 35.00 & 33.43 \\
Qwen2.5-Coder-7B-Instruct & 75.00 & 51.00 & 33.00 & 24.00 & 38.00 & 51.00 & 42.00 & 44.86 & 63.00 & 50.00 & 20.00 & 19.00 & 31.00 & 40.00 & 35.00 & 36.86 \\
Qwen2.5-14B-Instruct & 83.00 & 67.00 & 27.00 & 24.00 & 21.00 & 59.00 & 8.00 & 41.29 & 50.00 & 56.00 & 25.00 & 15.00 & 14.00 & 28.00 & 23.00 & 30.14 \\
Qwen2.5-72B-Instruct & 85.00 & 57.00 & 16.00 & 26.00 & 18.00 & 66.00 & 4.00 & 38.86 & 69.00 & 51.00 & 14.00 & 13.00 & 13.13 & 54.00 & 4.00 & 31.19 \\
Gemma2-2B-It & OOC & OOC & OOC & OOC & OOC & OOC & OOC & OOC & OOC & OOC & OOC & OOC & OOC & OOC & OOC & OOC \\
Gemma2-9B-It & OOC & OOC & OOC & OOC & OOC & OOC & OOC & OOC & OOC & OOC & OOC & OOC & OOC & OOC & OOC & OOC \\
Gemma2-27B-It & OOC & OOC & OOC & OOC & OOC & OOC & OOC & OOC & OOC & OOC & OOC & OOC & OOC & OOC & OOC & OOC \\
DeepSeek-V3 & 93.00 & 90.00 & 58.76 & 64.95 & 69.07 & 80.81 & 51.00 & 72.61 & 93.00 & 87.88 & 44.90 & 47.31 & 70.53 & 83.51 & 52.00 & 68.62 \\
\hline
\rowcolor{lightgray}
\multicolumn{17}{c}{Table Specific Models} \\
\hline
TableGPT2-7B & 67.00 & 52.00 & 32.00 & 19.00 & 31.00 & 36.00 & 40.00 & 39.57 & 43.00 & 46.00 & 24.00 & 19.00 & 31.00 & 19.00 & 38.00 & 31.43 \\
TableLlama & OOC & OOC & OOC & OOC & OOC & OOC & OOC & OOC & OOC & OOC & OOC & OOC & OOC & OOC & OOC & OOC \\
\hline
\rowcolor{lightgray}
\multicolumn{17}{c}{Close-Source Models} \\
\hline
GPT-4o-mini & 82.00 & 68.00 & 44.00 & 38.00 & 59.00 & 63.00 & 38.00 & 56.00 & 74.00 & 73.00 & 31.00 & 34.00 & 47.00 & 58.00 & 36.73 & 50.57 \\
GPT-4o & 88.00 & 93.00 & 56.00 & 54.00 & 60.00 & 81.00 & 50.00 & 68.86 & 90.62 & 85.42 & 46.74 & 32.63 & 54.95 & 83.33 & 47.78 & 63.41 \\
GPT-o1-mini & 79.80 & 77.78 & 78.00 & 78.79 & 64.00 & 68.00 & 50.00 & 70.88 & 73.00 & 77.55 & 39.80 & 42.55 & 60.42 & 57.14 & 52.00 & 57.60 \\
GPT-o3-mini & 89.00 & 87.00 & 91.00 & 96.00 & 89.00 & 87.00 & 63.00 & 86.00 & 80.00 & 82.83 & 78.00 & 78.00 & 75.00 & 74.00 & 53.54 & 74.50 \\
GPT-5.1 (Direct Prompt) & 95.00 & 94.00 & 88.00 & 78.00 & 82.00 & 94.95 & 54.00 & 83.69 & 94.90 & 93.88 & 57.58 & 57.14 & 72.92 & 91.00 & 53.54 & 74.42 \\
GPT-5.1 (Agent) & 94.00 & 94.00 & 87.00 & 76.77 & 81.00 & 92.00 & 56.00 & 82.98 & 95.00 & 95.00 & 62.63 & 60.00 & 78.00 & 87.00 & 60.53 & 77.48 \\
Gemini-3.1-Pro-Preview & 95.00 & 96.00 & 99.00 & 100.00 & 96.00 & 98.00 & 96.00 & 97.14 & 95.00 & 97.00 & 98.00 & 100.00 & 94.00 & 100.00 & 98.00 & 97.43 \\
 \hline
\rowcolor{lightgray}
\multicolumn{17}{c}{Reasoning Models} \\
\hline
DeepSeek-R1-Distill-Qwen-7B & 25.00 & 27.00 & 9.00 & 8.00 & 11.00 & 18.00 & 23.00 & 17.29 & NFI & NFI & NFI & NFI & NFI & NFI & NFI & NFI \\
DeepSeek-R1-Distill-Qwen-14B & 78.00 & 49.00 & 18.00 & 12.00 & 8.00 & 42.00 & 15.00 & 31.71 & 57.00 & 37.00 & 20.00 & 14.00 & 15.00 & 24.00 & 31.00 & 28.29 \\
DeepSeek-R1 & 94.00 & 98.00 & 94.00 & 94.95 & 87.00 & 96.00 & 63.00 & 89.56 & 92.63 & 91.11 & 75.00 & 81.61 & 80.22 & 92.31 & 58.33 & 81.50 \\
QwQ-32B-Preview & 3.00 & 2.00 & 1.00 & 0.00 & 2.00 & 2.00 & 9.00 & 2.71 & OOC & OOC & OOC & OOC & OOC & OOC & OOC & OOC \\
 \hline
\end{tabular}%
}
\label{tab:broader_test}
\vspace{-1em}
\end{table*}

\begin{figure*}[h!]
\centering
\includegraphics[width=0.95\textwidth]{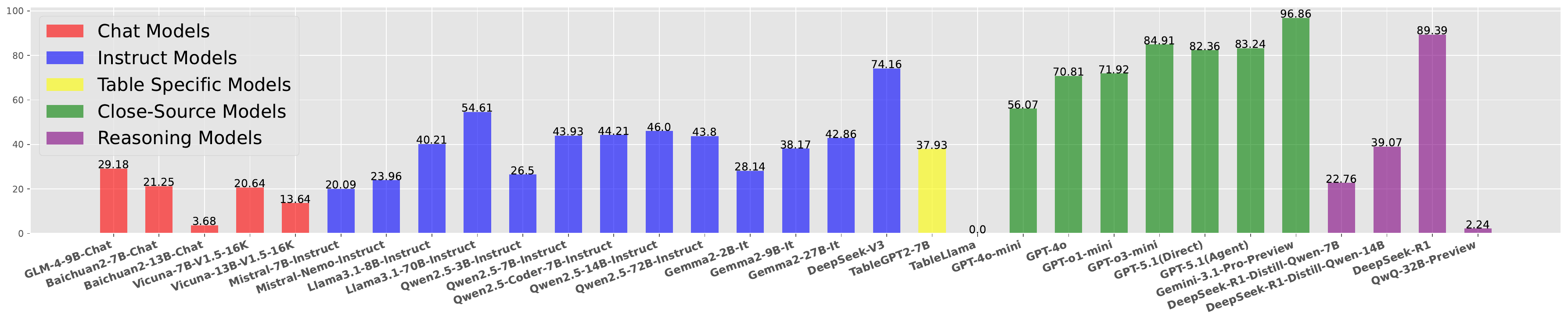}
\vspace{-1.0em}

\caption{The overall accuracy of all models.}
\label{fig:bar}
\end{figure*}

We first examine these results from a category-wise perspective by comparing the performance of chat, instruct, domain-specific, and reasoning models.

\noindent\textbf{Chat LLM Performance.} We find that \textit{chat-oriented LLMs generally perform poorly on our benchmark due to lack of instruction following ability}. Chat Models except \textsc{GLM4-9B-Chat} get lower than $25\%$ overall accuracy. These LLMs struggle to properly follow the instructions, despite the explicit clarification that there is only one correct choice and that the answer must adhere to a specific format. Instead, chat models often produce responses such as ``\texttt{I don’t know}'', ``\texttt{None of the above}'', or provide multiple answers in a format that deviates from the expected structure. This behavior likely stems from their design, which prioritizes conversational fluency over strict adherence to task-specific instructions. 
Interestingly, we also find that \textit{the overall accuracy of chat models tends to decline as their scale increases}. A manual inspection using the airline database revealed distinct failure patterns for different models. For instance, larger versions of \textsc{Baichuan2} frequently exhibited repetitive output (e.g., ``\texttt{-338.166666666666...}'' repeated many times), whereas smaller versions, while often failing to follow instructions, still managed to produce readable and structured responses. For \textsc{Vicuna} series, smaller models occasionally generated multiple-choice answers (e.g., ``\texttt{Answer: C/D}''), where our regular expression matched one option (e.g., ``\texttt{C}''), giving them a chance to answer correctly. In contrast, larger models often produced verbose analyses or answers like ``\texttt{None of the above}'', which made it impossible for our evaluation to match the correct answer.

\noindent\textbf{Instruct LLM Performance.} On the other hand, we find that \textit{instruct models demonstrated superior adherence to instructions to generally perform better.} In fact, most instruct LLMs achieve overall accuracy of $25\%$ or higher. As expected, larger instruct models generally performed better, showing a clear trend of improved accuracy with scale. Notably, coding LLMs within this category performed slightly better than their standard instruction-following counterparts. However, the performance gap was relatively marginal, suggesting that this observation may not generalize across all tasks or datasets.
Notice that two models deviate from the aforementioned observation --- \textsc{Mistral} and \textsc{Qwen2.5}. For \textsc{Mistral}, our manual analysis reveals that, despite being an instruction-tuned model, it frequently generates responses such as ``\texttt{None of the above}''. We hypothesize that its ability to follow instructions diminishes when dealing with long-context tables. For \textsc{Qwen2.5}, the largest version does not show a significant improvement in overall accuracy. Our manual review of the generated outputs suggests that this is due to its tendency to produce overly detailed analyses, often repeating information that has already been provided. This verbosity causes it to frequently exceed the token limit and finally prevents it from providing a concrete final answer.

\noindent \textbf{Domain-Specific Tabular LLMs.} Surprising, we find a fact that \textit{ 
LLMs that are specialized and fine-tuned for relational tasks, such as \textsc{TableLlama} and \textsc{TableGPT2}, did not perform as well as anticipated.} \textsc{TableLlama}, in particular, failed to follow the required answer format {consistently} entirely. \textsc{TableGPT2}, while capable of adhering to the format more consistently, delivered only {moderate} average performance. We suspect that the specialization of these models may inadvertently narrow their focus and make them less adaptable to the variations in question types or formats that were not adequately represented in their continuous pre-training corpus. Furthermore, despite being tailored for table-based tasks, these LLMs may not optimally balance between relational data generation accuracy and the flexibility required for general QA tasks. This suggests a potential systematic misalignment between training objectives and the evaluation criteria introduced in our benchmarks.

\noindent \textbf{Reasoning LLM Performance.} Our experimental results reveal \textit{notable disparities in the performance of reasoning models on our benchmark}. While recent popular reasoning models are generally expected to excel in logical and analytical tasks, our findings suggest otherwise for distilled models. Specifically, distilled versions fail to surpass their original counterparts, like the distilled \textsc{Qwen} models, which perform worse than the original ones. Moreover, distillation negatively impacts small models' ability to handle long contexts; for instance, \textsc{Qwen2.5-7B-Instruct} successfully follows instructions in a 64K scale, whereas its distilled version fails. Additionally, \textsc{QwQ-32B-Preview} exhibits anomalous behavior, frequently generating repetitive and meaningless tokens, a pattern also observed in \textsc{DeepSeek-R1-Distill-Qwen-7B} at the 64K scale. Overall, reasoning models display significant performance variation, with some achieving outstanding results. Notably, \textsc{DeepSeek-R1} demonstrates strong performance across all evaluated scales.

Beyond the comparison above, we examine three additional factors across model families: agentic reasoning, answer-option removal, and context length.

\noindent \textbf{Agentic Performance.}
We further compare \textsc{GPT-5.1} under two settings: direct prompting and an agentic setting, where the model can conduct multi-round planning and invoke tools to refine its intermediate analysis {iteratively} before producing the final answer. Compared to direct prompting, the agentic setup yields only limited improvements {overall}. Notably, the gains become more evident at the 64K context length, yet the absolute accuracy remains far from satisfactory, suggesting that tool-augmented reasoning still struggles on multi-table QA at scale. These results also highlight that context length continues to be a factor shaping agentic performance.

\noindent \textbf{Supplementary Open-Ended Pilot.} As a complementary check beyond the MCQ setting, we additionally evaluate an aligned open-ended value-generation pilot. On the paired 700-question subset with \textsc{GPT-4o-mini} at 8K, accuracy drops from 60.86\% in the MCQ setting to 33.71\% in the open-ended setting. The largest degradation appears on \textit{count}, \textit{average}, \textit{sum}, and \textit{correlation}, indicating that removing answer options makes numerical and statistical reasoning substantially harder. Detailed prompting, evaluation rules, and per-type results are provided in Appendix~\S{II}.

\noindent \textbf{Impact of Context Length.} We analyze how table context length affects representative models across model families, and analyze their performance across different question subcategories as the context length changes.
Our experiments show that \textit{the performance of most models generally decreases as the context length increases}. Although \textsc{Gemini-3.1-Pro-Preview} remains highly stable across context lengths and several models show slight improvements in specific settings, the broader pattern indicates that handling longer tabular contexts remains challenging for LLMs. We enumerate the effect of context length across task categories below:

\begin{itemize}[topsep=5pt, leftmargin=1em]
    \item \textit{The performance of lookup tasks declines relatively slowly with increasing context length}. Since these tasks often require retrieving only a single or a few items from the table, they remain manageable even with longer contexts.
    
    \item \textit{LLMs experience more pronounced performance drops in this aggregation task}, as aggregation tasks typically involve explicit computations over multiple elements. Notably, most models perform better in answering the \textit{average} questions than in the \textit{sum} questions. This disparity arises because estimating a value close to the average is more intuitive, potentially through an informed educated guess, whereas summing multiple elements accurately requires more robust analytical and mathematical abilities.

    \item For complex calculation questions, the impact of longer contexts depends on the specific subcategory. \textit{The composite comparison task reserves relatively good performance}, as some question instances only require retrieving values and performing simple comparisons. The \textit{correlation calculation tasks encounter a significant performance drop}. Perhaps because it requires both complex numerical computations and logical reasoning, making it harder for models to handle, especially with larger tables.
\end{itemize}

\begin{figure}[h!]
\centering
\includegraphics[width=0.5\textwidth]{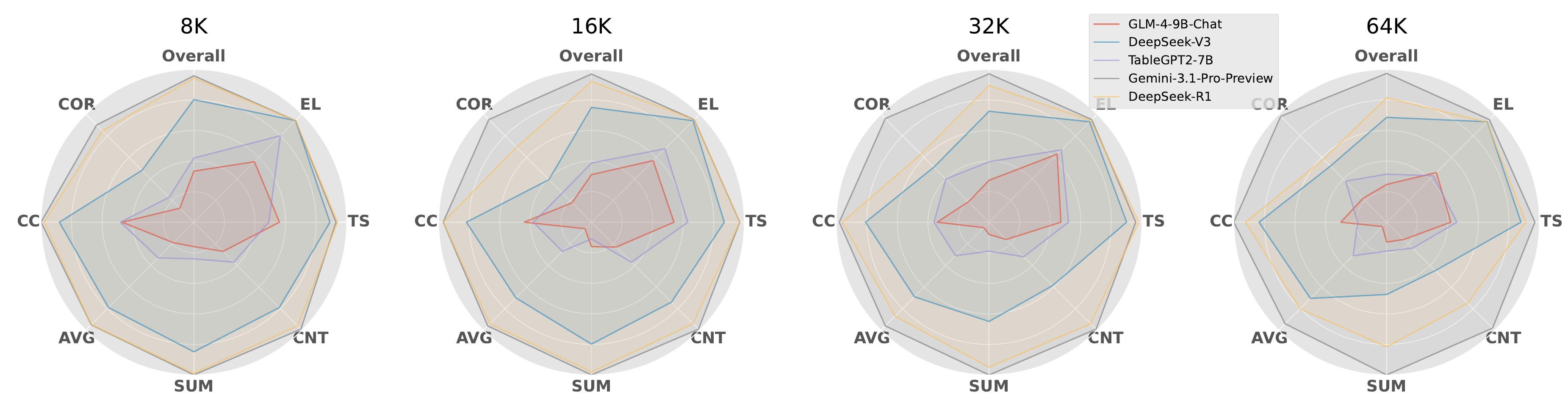}

\caption{The accuracy distribution for subcategories in different context lengths.}

\label{fig:circle}
\end{figure}

In summary, \textit{as the context length increases, most models struggle with all but the simplest questions}. Fig.~\ref{fig:circle} illustrates this overall pattern while also showing that strong frontier models such as \textsc{Gemini-3.1-Pro-Preview} can remain robust. For most models, even in relatively simpler subcategories, performance steadily declines as the table context grows. These findings highlight the persistent difficulties in handling complex reasoning tasks in longer context tables.

\begin{tcolorbox}[colback=blue!5!white,colframe=blue!75!black]
  \textbf{Answer to Question (\underline{iii})}:
\textit{Our comprehensive evaluation highlights {several} interesting observations: instruct-tuned LLMs exhibit significantly better task performance compared to chat-oriented LLMs. Specialized tabular LLMs demonstrate limited flexibility, underperforming relative to expectations. Reasoning LLMs can exhibit strong performance (i.e., \textsc{Deepseek-R1}), whereas distillation may negatively impact accuracy and long-context handling. Moreover, longer context lengths usually challenge LLMs, with substantial performance drops in aggregation and complex calculation tasks. A supplementary aligned open-ended pilot further shows that removing answer options substantially increases difficulty, especially for numerical and statistical questions.}
\end{tcolorbox}

\subsection{Sampling and Symbolic Extension}

\begin{figure}[h!]
\centering
\includegraphics[width=0.5\textwidth]{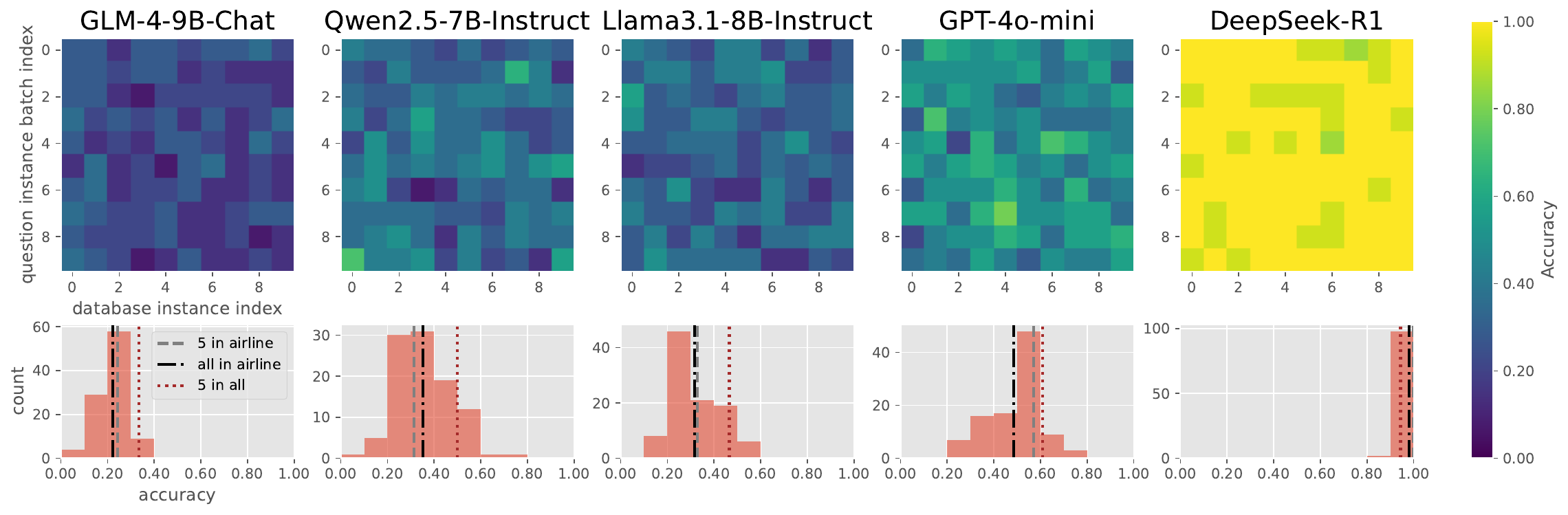}
\vspace{-2em}
\caption{The accuracy distribution of question instances in context length 8K, airline database between five models.}

\label{fig:sensitive}
\end{figure}

To examine the impact of sampling and symbolic extensions on model performance, Experiment 4 focuses on the \texttt{airline} database at 8K and evaluates 1,400 generated questions from this setting alone.
To analyze the results, we created a heatmap to illustrate the accuracy of each question instance batch for all models and a histogram to show the accuracy distribution across batches for each model. We also compared the average accuracy between three sets of batches: all batches from the \texttt{airline} database (all in airline), 5 batches previously used from \texttt{airline} (5 in airline), and 5 batches previously used across all databases (5 in all). These visualizations are shown in Fig.~\ref{fig:sensitive}.

\noindent \textbf{Results and Detailed Analysis.}
We summarize our detailed analysis of the influence of the sampling and symbolic extension. \textit{The heatmap results highlight that question difficulty varies across batches and models}. For example, a batch in the lower-left corner of the heatmap appears easier for the \textsc{Qwen2.5-7B-Instruct}, achieving high accuracy, while the same batch presents an average difficulty for other models, indicated by a consistent green color. This finding suggests that different models have distinct preferences or tendencies when handling certain question instances. Therefore, incorporating sampling and symbolic extensions helps reduce evaluation instability by enlarging and diversifying the generated question pool. Furthermore, the histogram shows a detailed distribution of the accuracy, which indicates that \textit{if the quantity of the generated question is small, we might evaluate model performance imprecisely.} This underscores the value of sampling and symbolic extensions in ensuring a more balanced and thorough evaluation of models.
Lastly, we compare the sampling results at the single-database and pooled-benchmark levels in a more systematic way. On the single \texttt{airline} database, 5-batch estimates still show noticeable variation, especially for closely matched models; when aggregated over all ten databases, the same sampling design yields more stable model rankings and only modest pooled-score uncertainty. Appendix~\S{II} reports the detailed stability statistics, and these results indicate that sampling and symbolic extension are most useful for supporting stable benchmark-level comparisons rather than replacing exact single-database estimates.

\begin{tcolorbox}[colback=blue!5!white,colframe=blue!75!black]
  \textbf{Answer to Question (\underline{iv})}:
\textit{The sampling mechanism combined with our sampling and symbolic extension improves benchmark stability by enlarging and diversifying the generated question pool. At the single-database level, sampled batches can still vary in difficulty, but at the benchmark level the same design supports more stable model rankings and pooled performance trends across sampled evaluations.}
\end{tcolorbox}

\subsection{Direct LLM Prompt vs. Text2SQL}

To systematically compare direct LLM prompting with LLM-based Text2SQL, we conducted experiments across multiple context lengths using the same questions and database instances as in \textbf{Experiment 3}, covering over 2,800 questions in total. In the Text2SQL setup, LLMs are provided with database schemas and instructed to generate executable SQL queries, using a prompt adapted from \textsc{Arctic-Text2SQL-R1}~\cite{yao2025arctic}. The generated SQL is executed on the corresponding sampled database, and the returned answer is evaluated by the type-aware answer-layer protocol described in Section~\S\ref{sec:eval_design} and Appendix~\S{II}. We tested six models: \textsc{GLM-4-9B-Chat}, \textsc{Qwen2.5-Coder-7B-Instruct}, \textsc{Llama3.1-8B-Instruct}, \textsc{Arctic-Text2SQL-R1}, \textsc{GPT-4o-mini}, and \textsc{DeepSeek-R1}. Among them, \textsc{Arctic-Text2SQL-R1} is a specialized Text2SQL model and ranks top on BIRD-Bench among comparable open models~\cite{bird_leaderboard}.

\begin{table*}[htbp]
\centering
\caption{Text2SQL performance on the multi-table TQA-Bench under the answer-layer evaluation protocol.}
\resizebox{\textwidth}{!}{%
\begin{tabular}{ l|ccccccc|c|ccccccc|c  }
 \hline
 & \multicolumn{8}{|c|}{\textbf{8K}} & \multicolumn{8}{|c}{\textbf{16K}} \\
 \hline
 \textbf{Model} & \textbf{EL} & \textbf{TS} & \textbf{CNT} & \textbf{SUM} & \textbf{AVG} & \textbf{CC} & \textbf{COR} & \textbf{Overall} & \textbf{EL} & \textbf{TS} & \textbf{CNT} & \textbf{SUM} & \textbf{AVG} & \textbf{CC} & \textbf{COR} & \textbf{Overall} \\
\hline
GLM-4-9B-Chat & 54.00 & 27.00 & 46.00 & 46.00 & 27.00 & 26.00 & 2.00 & 32.57 & 49.00 & 21.00 & 47.00 & 32.00 & 29.00 & 26.00 & 1.00 & 29.29 \\
Qwen2.5-Coder-7B-Instruct & 31.00 & 20.00 & 32.00 & 36.00 & 26.00 & 21.00 & 1.00 & 23.86 & 40.00 & 16.00 & 34.00 & 37.00 & 36.00 & 17.00 & 1.00 & 25.86 \\
Llama3.1-8B-Instruct & 39.00 & 11.00 & 33.00 & 29.00 & 33.00 & 15.00 & 0.00 & 22.86 & 33.00 & 12.00 & 28.00 & 34.00 & 21.00 & 8.00 & 0.00 & 19.43 \\
Arctic-Text2SQL-R1-7B & 74.00 & 50.00 & 78.00 & 67.00 & 74.00 & 51.00 & 13.00 & 58.14 & 75.00 & 43.00 & 72.00 & 68.00 & 72.00 & 49.00 & 11.00 & 55.71 \\
GPT-4o-mini & 74.00 & 55.00 & 77.00 & 64.00 & 73.00 & 50.00 & 2.00 & 56.43 & 78.00 & 45.00 & 72.00 & 64.00 & 72.00 & 48.00 & 0.00 & 54.14 \\
DeepSeek-R1 & 85.00 & 73.00 & 80.00 & 83.00 & 88.00 & 66.00 & 90.00 & 80.71 & 87.00 & 69.00 & 78.00 & 80.00 & 84.00 & 71.00 & 87.00 & 79.43 \\
 \hline
 & \multicolumn{8}{|c|}{\textbf{32K}} & \multicolumn{8}{|c}{\textbf{64K}} \\
 \hline
GLM-4-9B-Chat & 53.00 & 30.00 & 48.00 & 41.00 & 25.00 & 28.00 & 1.00 & 32.29 & 50.00 & 26.00 & 44.00 & 35.00 & 29.00 & 42.00 & 0.00 & 32.29 \\
Qwen2.5-Coder-7B-Instruct & 27.00 & 24.00 & 38.00 & 26.00 & 25.00 & 14.00 & 4.00 & 22.57 & 39.00 & 22.00 & 30.00 & 25.00 & 23.00 & 25.00 & 0.00 & 23.43 \\
Llama3.1-8B-Instruct & 42.00 & 12.00 & 33.00 & 34.00 & 22.00 & 10.00 & 0.00 & 21.86 & 29.00 & 11.00 & 28.00 & 30.00 & 18.00 & 14.00 & 0.00 & 18.57 \\
Arctic-Text2SQL-R1-7B & 75.00 & 48.00 & 74.00 & 65.00 & 68.00 & 50.00 & 11.00 & 55.86 & 74.00 & 43.00 & 61.00 & 60.00 & 69.00 & 51.00 & 10.00 & 52.57 \\
GPT-4o-mini & 73.00 & 54.00 & 74.00 & 63.00 & 69.00 & 50.00 & 1.00 & 54.86 & 74.00 & 49.00 & 65.00 & 56.00 & 62.00 & 44.00 & 0.00 & 50.00 \\
DeepSeek-R1 & 84.00 & 74.00 & 73.00 & 80.00 & 81.00 & 66.00 & 89.00 & 78.14 & 84.00 & 65.00 & 66.00 & 76.00 & 80.00 & 70.00 & 86.00 & 75.29 \\
 \hline
\end{tabular}%
}
\label{tab:sqltest}
\vspace{-1em}
\end{table*}

\noindent \textbf{Results and Detailed Analysis.} 
The Text2SQL results under the answer-layer evaluation protocol are listed in TABLE~\ref{tab:sqltest}. Compared with the comprehensive direct-prompt results in TABLE~\ref{tab:broader_test}, they reveal three main observations. \underline{First}, \textit{Text2SQL remains substantially less sensitive to context length than direct prompting}. For example, \textsc{DeepSeek-R1} drops only from 80.71\% at 8K to 75.29\% at 64K, and \textsc{Arctic-Text2SQL-R1-7B} drops from 58.14\% to 52.57\%, showing that schema-only prompting is relatively robust to longer serialized databases. 
\underline{Second}, \textit{the strongest Text2SQL model shows broad advantages across the benchmark}.
Under the answer-layer protocol, \textsc{DeepSeek-R1} remains the strongest model, followed by \textsc{Arctic-Text2SQL-R1-7B} and \textsc{GPT-4o-mini}; its advantage is especially pronounced on COR, as shown in TABLE~\ref{tab:sqltest}.
\underline{Third}, \textit{complex analytical SQL remains the main bottleneck}. The contrastive analysis in Appendix~\S{III} shows two complementary phenomena: \textsc{DeepSeek-R1} more often composes scalar SQL that stays close to the intended analytical semantics, while \textsc{Arctic-Text2SQL-R1-7B} still fails mainly because it reformulates Composite Comparison questions into counts or comparisons and emits proxy outputs or incomplete Pearson-style SQL for Correlation. Therefore, Text2SQL remains complementary to direct prompting, while complex analytical SQL continues to be the main limitation on the hardest workloads in our setting.

\begin{tcolorbox}[colback=blue!5!white,colframe=blue!75!black]
  \textbf{Answer to Question (\underline{v})}:
\textit{Under the type-aware answer-layer evaluation, LLM-based Text2SQL methods remain relatively stable across varying context lengths and therefore complement direct prompting approaches. However, complex analytical tasks are still the main bottleneck: strong reasoning models more often stay close to the intended numerical semantics, whereas specialized Text2SQL models still fail mainly on composite-comparison formulation and statistical SQL construction in more challenging cases.}
\end{tcolorbox}

\section{Related Work}

\noindent\textbf{Table QA.}
Question answering (QA) over relational databases has long been central to natural language processing and data management~\cite{jin2022survey}. Given a user query, table QA seeks accurate answers via table understanding and reasoning~\cite{pal2023multitabqa, zhang2024reactable,zhu2024autotqa}. Methods are commonly grouped into two classes: (i) \emph{Text2SQL}, which translates natural language into executable SQL~\cite{liu2021tapex,fu2023catsql,gu2023few,li2024codes,fan2024combining,zhang2024reactable,katsogiannis2023survey}; and (ii) \emph{end-to-end} models that process the question with a serialized table to directly produce an answer~\cite{fetaqa,multihiertt,hitab,ottqa,hybridqa}. Representative E2E systems include \textsc{Table-BERT}~\cite{chentabfact}, which converts tables into coherent text for downstream processing; \textsc{TaPas}~\cite{herzig2020tapas}, which encodes tables within BERT; and PASTA~\cite{gu2022pasta}, which pre-trains on cloze-style sentence–table tasks using WikiTables. Multi-table QA has been explored by \textsc{MultiTabQA}~\cite{pal2023multitabqa}, while AutoTQA~\cite{zhu2024autotqa} uses multi-agent LLMs for conversational solving. 
Our benchmark focuses on comprehensively evaluating techniques in the second class with reliable, comparable results.

\noindent\textbf{Assessment of LLM over data management tasks.} LLMs have demonstrated remarkable capabilities, offering an opportunity to build a new wave of innovative AI applications~\cite{bommasani2021opportunities}. In the data management community, LLM has also reshaped the paradigm of design and implementation of many data management tasks~\cite{biswal2024text2sql,chentablerag,patel2024lotus,wornow2024automating} including data integration~\cite{huo2024zeroea,dohmen2024schemapile}, database system tuning~\cite{giannakouris2024demonstrating}, query optimization
~\cite{liu2024optimizing}, table summarization~\cite{liu2022long}, table formatting~\cite{singh2023format5} etc. Various domain-specific LLMs for tabular data have also been released. For example, \textsc{TableLlama}~\cite{zhang2023tablellama} is fine-tuned on the TableInstruct dataset, which comprises 2.6 million table-based tasks to process 8 in-domain table-related tasks. 
\textsc{TableGPT}~\cite{zha2023tablegpt,su2024tablegpt2} aims to unify tables, natural language, and commands into a single LLM to enable seamless interaction with tables, supporting functionalities such as question answering, data manipulation, visualization, and analysis report generation. To evaluate the performance of LLM over data management tasks, different benchmarks have been released, including benchmarks for Text2SQL~\cite{yu2018spider,lei2024spider,gao2024text}, relational structural understanding capabilities~\cite{sui2024table}, and table QA~\cite{chentabfact,lei2023tableqakit,wu2024tablebench}. Our benchmark is designed as a new benchmark for multi-table QA with varied relational data contexts and symbolic extensions to access the reasoning capabilities of LLMs.

\noindent\textbf{Table-Specific Reasoning Mechanisms.}
Beyond domain-specific fine-tuning, recent work has also explored \emph{table-specific reasoning mechanisms}. \textsc{Chain-of-Table}~\cite{wang2024chain} treats intermediate tabular states as structured thoughts by iteratively updating the table during reasoning; \textsc{Tree-of-Table}~\cite{ji2024tree} first condenses and decomposes large tables and then performs hierarchical tree-structured reasoning; and \textsc{Chain-of-Query}~\cite{sui2025chain} uses SQL-aided multi-agent collaboration with clause-by-clause query construction and a division between SQL-based mechanical reasoning and LLM-based logical inference. These approaches are complementary to TQA-Bench: rather than proposing another reasoning mechanism, our goal is to provide a controlled analytical multi-table benchmark that can reveal how such strategies behave under long-context, multi-table conditions.

\section{Conclusion}
In this paper, we introduce TQA-Bench, a new multi-table QA benchmark designed to rigorously evaluate the capabilities of LLMs in processing complex, relational data across multiple tables. 
Our benchmark applies diverse relational database instances drawn from real-world public datasets, a flexible sampling mechanism that allows for the creation of tasks with varying context lengths from 8K to 64K tokens, and the integration of symbolic extensions to test higher-order reasoning capabilities.
Through systematic evaluations involving both open-source and closed-source LLMs, with scales ranging from 2 billion to 671 billion parameters, our findings highlight the variable performance of these models under complex multi-table QA scenarios. We expect that TQA-Bench can serve as a pivotal step toward realizing the full potential of LLMs in the analysis of complex tabular data.
Our Text2SQL study further evaluates numerical analytical SQL with a type-aware answer-layer protocol. Under this protocol, most models still struggle with complex analytical tasks, especially Composite Comparison and Correlation.

At the same time, the current benchmark is intentionally scoped to analytical multi-table QA: it emphasizes retrieval, aggregation, comparison, and correlation over realistic relational schemas, but does not yet cover broader table-understanding tasks such as summarization, induction, or open-ended information extraction. MCQ therefore remains the primary evaluation format for scalable and low-ambiguity benchmarking, while the supplementary open-ended pilot should be viewed as an initial check rather than a second full evaluation track. Similarly, our sampling results are strongest at the pooled benchmark level, where rankings remain stable across databases, rather than as a universal guarantee that a small batch budget precisely substitutes for every individual database. Finally, although TQA-Bench already spans diverse real-world databases and a broad model set, extending the benchmark to deeper schema structures, broader task families, newer frontier models, and table-specific reasoning mechanisms such as table-state evolution, hierarchical decomposition, and SQL-aided collaboration remains an important direction for future work.


\bibliographystyle{IEEEtran}
\bibliography{refs}

\clearpage
\section*{Biography}

\begin{IEEEbiographynophoto}{Zipeng Qiu}
Zipeng Qiu received the B.S. degree from Fudan University, Shanghai, China. He is currently a Ph.D. student in the Department of Computer Science and Engineering at the Hong Kong University of Science and Technology (HKUST), Hong Kong, China, supervised by Prof. Binhang Yuan. His research interests include data management, large language model systems, and LLM-based question answering over structured data.
\end{IEEEbiographynophoto}

\begin{IEEEbiographynophoto}{Chenyue Li}
Chenyue Li received the Honours B.S. degree in computer science from the University of Toronto, Toronto, Canada. He is currently a Ph.D. student in the Department of Computer Science and Engineering at HKUST, Hong Kong, China, supervised by Prof. Binhang Yuan. His research interests include AI for science, large language models, database systems, distributed systems, and full-stack development.
\end{IEEEbiographynophoto}

\begin{IEEEbiographynophoto}{You Peng}
You Peng received the B.Sc. degrees in data science and statistics from the University of Toronto, Toronto, Canada. He is currently a Ph.D. student in the Department of Computer Science and Engineering at HKUST, Hong Kong, China, supervised by Prof. Binhang Yuan. His research interests include database-related LLM agents, machine learning systems, and data science.
\end{IEEEbiographynophoto}

\begin{IEEEbiographynophoto}{Guangxin He}
Guangxin He received the B.S. degree from the University of Chinese Academy of Sciences, Beijing, China, and the master's degree from the Institute of Computing Technology, Chinese Academy of Sciences, Beijing, China. He is currently a Ph.D. student in the Department of Computer Science and Engineering at HKUST, Hong Kong, China, supervised by Prof. Binhang Yuan. His research interests include LLM systems and retrieval-augmented generation for large language models.
\end{IEEEbiographynophoto}

\begin{IEEEbiographynophoto}{Binhang Yuan}
Binhang Yuan received the B.S. degree in computer science from Fudan University, Shanghai, China, and the M.S. and Ph.D. degrees in computer science from Rice University, Houston, TX, USA. He is currently an Assistant Professor in the Department of Computer Science and Engineering at HKUST, Hong Kong, China, where he leads the Relaxed System Lab. Before joining HKUST, he was a Postdoctoral Researcher at ETH Zurich, Zurich, Switzerland. His research interests include data management systems for machine learning and distributed and decentralized machine learning systems.
\end{IEEEbiographynophoto}

\begin{IEEEbiographynophoto}{Chen Wang}
Chen Wang is a Research Associate Professor and CTO at the National Engineering Research Center for Big Data Software, Tsinghua University, Beijing, China, where he also serves as Director of the Big Data Research Department at the Energy Internet Research Institute. Before joining Tsinghua University, he was a Research Staff Member and Senior Manager in the Information Management Research Department at IBM Research China. His research interests include database systems, data governance, time-series data management, and industrial big data applications. He is a founding member and PMC member of the Apache IoTDB project.
\end{IEEEbiographynophoto}

\end{document}